\documentclass[10pt,twocolumn,letterpaper]{article}
\PassOptionsToPackage{table,xcdraw,dvipsnames}{xcolor}

\usepackage{times}
\usepackage{xspace}
\usepackage{xcolor}
\usepackage{graphicx}
\usepackage{amsmath}
\usepackage{amssymb}
\usepackage{booktabs}
\usepackage[numbers,sort&compress]{natbib}
\usepackage[format=plain,labelformat=simple,labelsep=period,font=small,compatibility=false]{caption}
\usepackage[font=footnotesize,skip=3pt,subrefformat=parens]{subcaption}
\usepackage[hyphens]{url}
\usepackage{etoolbox}
\usepackage{algorithm}
\usepackage{algorithmic}
\usepackage{array}
\usepackage{pifont}
\usepackage{placeins}
\usepackage{float}
\usepackage{wasysym}
\usepackage{enumitem}
\usepackage{wrapfig}
\usepackage{tikz}

\makeatletter
\def\onedot{\futurelet\@let@token\@onedot}
\def\@onedot{\ifx\@let@token.\else.\null\fi\xspace}

\def\etal{\emph{et al}\onedot}
\makeatother

\newcommand {\myvec}[1] {{\mbox{\boldmath $#1$}}}
\newcommand{\tauminpix}{\tau_{\mathrm{minpix}}}   
\newcommand{\tauIoU}{\tau_{\mathrm{IoU}}}         
\newcommand{\lconcept}{\lambda_{\mathrm{concept}}} 

\definecolor{linkblue}{rgb}{0.21,0.49,0.74}
\usepackage[pagebackref,breaklinks,colorlinks,allcolors=linkblue]{hyperref}

\DeclareMathOperator*{\softmax}{softmax}

\newcommand{\openingfigureblock}{%
  \vspace{-0.6em}
  \begin{center}
    \includegraphics[width=\linewidth]{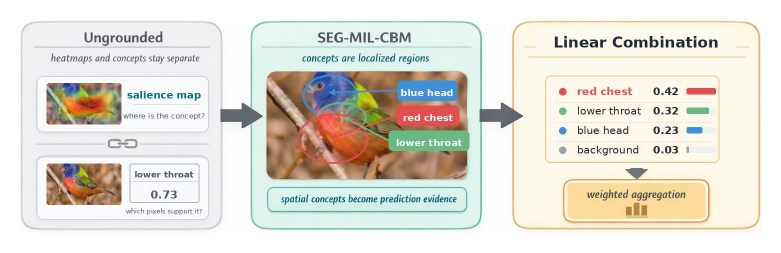}
    \captionof{figure}{
    \textbf{Spatially grounded concept reasoning.} Prior explanations either provide salience without semantic concepts or global concept scores without localized evidence. SEG-MIL-CBM grounds concept evidence in image regions and uses the same attention-weighted region--concept contributions both to form the prediction and to explain it.
    }
    \label{fig:opening_figure}
  \end{center}
  \vspace{0.4em}
}
\makeatletter
\renewcommand{\@maketitle}{%
  \vskip .375in
  \begin{center}%
    {\Large \bf \@title \par}%
    \vskip 24pt%
    {\large
      \begin{tabular}[t]{c}%
        \@author
      \end{tabular}\par}%
  \end{center}%
  \vspace{0.5em}%
}
\apptocmd{\@maketitle}{\openingfigureblock}{}{}
\makeatother
\title{Spatially Grounded Concept-Based Image Classification}

\author{Ran Eisenberg \quad Amit Rozner \quad Ethan Fetaya \quad Ofir Lindenbaum\\
Faculty of Engineering\\
Bar-Ilan University\\
Ramat Gan, 5290002, Israel\\
{\tt\small \{ran.eisenberg,amit.rozner,ethan.fetaya,ofir.lindenbaum\}@biu.ac.il}
}

\begin{document}
\maketitle

\begin{abstract}
Deep neural networks can achieve high accuracy while relying on evidence that is hard to inspect or misaligned with the intended task. Concept Bottleneck Models (CBMs) expose human-interpretable concepts, but most treat concepts as global attributes and do not show how localized evidence is aggregated into a decision. We propose \textbf{SEG-MIL-CBM}, a spatially grounded CBM that decomposes each image into concept-guided regions and classifies it by attention-based aggregation of segment-level concept evidence. The same segment evidence terms form the prediction and the explanation, exposing which regions and concepts support the predicted logit without a separate post-hoc attribution module. Among evaluated CBM-family baselines, SEG-MIL-CBM improves Waterbirds worst-group accuracy from $65.1\%$ to $72.0\%$, reaches $87.4\%$ worst-group accuracy on Pawrious, remains competitive on standard recognition, and attains the best CBM accuracy on CIFAR-100 ($85.3\%$). Segment-level faithfulness experiments on CUB further show that its learned segment ranking matches or improves over evaluated segment-ranking controls.
\end{abstract}

\section{Introduction}

Deep vision models are increasingly used in high-stakes settings, yet their predictions often depend on evidence that is difficult to inspect \cite{csahin2025unlocking,eke2025role}. This is especially problematic under spurious correlations: a Waterbirds classifier, for example, may rely on background rather than bird evidence \cite{sagawa2019distributionally}. Such shortcut reliance can harm worst-group accuracy and is difficult to diagnose from image-level predictions alone \cite{sun2023right,labonte2024group}.

Concept Bottleneck Models (CBMs) \cite{koh2020concept,oikarinenlabel,yuksekgonul2023posthoc,dcbm2025} provide a structured alternative by mapping images through human-interpretable concepts. However, most CBMs reason with global concept scores and do not identify which image regions support those concepts. Spatial CBMs such as SALF-CBM \cite{benou2025show} localize concept evidence, but map-level localization alone does not explicitly account for how region evidence is combined into the final logit. As a result, explanations can remain separate from the computation that produced the decision.

We argue that a stronger explanation should be an audit trail of the prediction itself. A model that assembles its class logit from localized concept evidence can report which region was used, which concepts were active there, and how much that region supported the decision. We introduce \textbf{SEG-MIL-CBM}, a concept-based vision model that decomposes an image into concept-guided regions using pretrained vision-language and segmentation models, treats the regions as instances in an attention-based multiple-instance learning (MIL) model, and reads explanations directly from the resulting segment-level logit decomposition.

The displayed explanation is not a post-hoc attribution on top of a separate classifier. Segments are ranked by the positive support they contribute to the predicted class, and the same local concept activations identify which concepts supply that support. A cosine alignment loss preserves the semantic concept interface without using explicit concept or group labels in the training objective.

Experimentally, SEG-MIL-CBM improves worst-group accuracy over evaluated CBM-family baselines on Waterbirds and Pawrious while remaining competitive on large-scale recognition. We also compare against non-CBM group-robust methods to make the accuracy-interpretability trade-off explicit: these methods can achieve higher worst-group accuracy, but do not provide spatially grounded concept accounting. On CUB, segment-level faithfulness experiments show that the learned segment ranking matches or improves over the strongest evaluated segment-ranking controls. To support reproducibility, we include code in the supplementary material.

Our key contributions are:
\begin{itemize}[noitemsep,topsep=0pt,leftmargin=*]
    \item We formulate spatial concept prediction as accountable aggregation: every displayed region score is read from the non-bias segment evidence term in the logit decomposition.
    \item We quantitatively validate this explanation-as-computation view through segment-level faithfulness: on CUB, the learned segment ranking matches or improves over the strongest evaluated deletion/insertion controls.
    \item We show that this interpretability-first design preserves competitive recognition accuracy, improves worst-group accuracy among evaluated CBM-family baselines, and exposes the Pareto trade-off against non-interpretable group-robust methods.
\end{itemize}

\section{Related Work}
\label{sec:related_work}

\textbf{Concept-based interpretability.} Concept Bottleneck Models (CBMs) \cite{koh2020concept} predict human-defined concepts before task classification, enabling concept-level reasoning but often requiring concept annotations. Post-hoc and label-free CBMs reduce this supervision burden by mapping pretrained features into concept spaces \cite{yuksekgonul2023posthoc,oikarinenlabel}, while adaptive and data-efficient variants extend the framework to foundation models \cite{choi2024adaptive,dcbm2025}. Spatial CBMs such as SALF-CBM \cite{benou2025show} localize concept evidence, but do not directly expose the region-level contributions that form the final prediction. A fuller criteria comparison is given in Appendix~\ref{appendix:group_robust_baselines}, Table~\ref{apptab:comparison_criteria_full}.

\textbf{Robustness and region aggregation.} Robustness methods for spurious correlations optimize group risk with group annotations (e.g., GroupDRO~\cite{sagawa2019distributionally}, DFR~\cite{kirichenko2022last}, DaC~\cite{Noohdani_2024_CVPR}) or infer/reweight groups without them (e.g., JTT~\cite{liu2021just}, EIIL~\cite{creager2021environment}, CnC~\cite{zhang2022correct}, AFR~\cite{qiu2023simple}, DISC~\cite{wu23disc}). These methods improve worst-group accuracy but generally do not provide concept-level or spatial explanations. SEG-MIL-CBM instead builds on open-vocabulary foundation models such as CLIP, GroundingDINO, and SAM \cite{radford2021learning,liu2024grounding,kirillov2023segment}, and on attention-based MIL \cite{dietterich1997solving,maron1998framework,ilse2018attention}, to aggregate semantically meaningful regions into accountable predictions.

\section{Background}
\label{sec:background}

\textbf{Foundation Models for Open-World Semantics:} Open-vocabulary foundation models such as CLIP~\cite{radford2021learning}, SAM~\cite{kirillov2023segment}, and Grounding DINO~\cite{liu2024grounding} enable zero-shot decomposition of images into semantically meaningful regions without exhaustive human supervision. We use them as off-the-shelf preprocessing tools to obtain concept-guided segments, rather than as end-to-end predictors.

\textbf{Multiple Instance Learning in Vision:} Multiple Instance Learning (MIL) provides a natural framework for settings where only bag-level labels are available and individual instances within a bag are unlabeled \cite{dietterich1997solving,maron1998framework}. In vision, MIL and related weakly supervised formulations aggregate information from regions, pixels, or patches and can localize evidence for image-level predictions. Attention-based MIL architectures \cite{ilse2018attention} expose which instances influence the prediction; our method adapts this idea to concept-guided regions whose contributions can be named in a semantic vocabulary.

\begin{table}[t]
\centering
\footnotesize
\setlength{\tabcolsep}{2pt}
\begin{tabular*}{\columnwidth}{@{\extracolsep{\fill}}lcc@{}}
\toprule
\textbf{Method} & \textbf{Spatial Localization} & \textbf{Spurious Mitigation} \\
\midrule
Post-hoc CBM \cite{yuksekgonul2023posthoc} & \ding{55} & \ding{51} \\
Label-Free-CBM \cite{oikarinenlabel} & \ding{55} & \ding{55} \\
LaBo \cite{yang2023language} & \ding{55} & \ding{55} \\
CDM \cite{panousis2023sparse} & \ding{55} & \ding{55} \\
DCLIP \cite{menon2022visual} & \ding{55} & \ding{55} \\
DN\text{-}CBM \cite{rao2024discover} & \ding{55} & \ding{55} \\
\midrule
SALF\text{-}CBM \cite{benou2025show} & \ding{51} & \ding{55} \\
DCBM \cite{dcbm2025} & \ding{51} & \ding{55} \\
\rowcolor[HTML]{C6EAD8} SEG-MIL-CBM (ours) & \ding{51} & \ding{51} \\
\bottomrule
\end{tabular*}
\caption{
    CBM-family comparison. Extended criteria are reported in Appendix~\ref{appendix:group_robust_baselines}, Table~\ref{apptab:comparison_criteria_full}.
}
\label{tab:comparison_criteria}
\end{table}

\section{Method}

\begin{figure*}[!t]
    \centering
    \includegraphics[width=\textwidth]{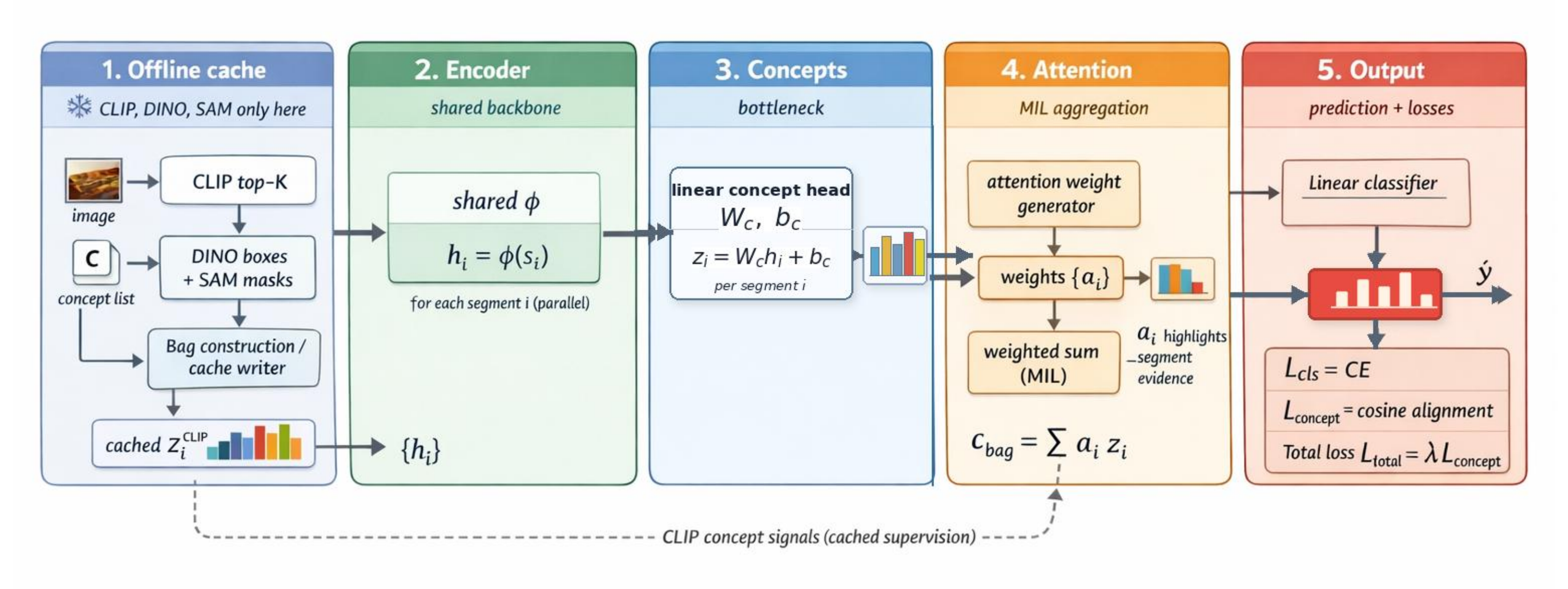}
    \caption{
        \textbf{Overview of the SEG-MIL-CBM training pipeline.} Offline preprocessing uses CLIP to select image-relevant concepts, GroundingDINO to localize them, and SAM to produce concept-guided masks, yielding a cached segment bag $\mathcal{B}_{\myvec{x}}=\{\myvec{s}_1,\dots,\myvec{s}_{N_s}\}$ and frozen CLIP concept-similarity targets $\myvec{c}^{\mathrm{CLIP}}_i$. The trainable model encodes each segment with a shared backbone $\myvec{h}_i=\phi(\myvec{s}_i)$, maps features through a linear concept head to obtain learned concept activations $\myvec{z}_i$, computes segment-class scores $g_{i,y}$ from $\myvec{w}_y^\top\myvec{z}_i+b_y$, and aggregates them with MIL attention. The final logit is $\ell_y=\sum_i \alpha_{i,y}g_{i,y}$, so each term $\alpha_{i,y}g_{i,y}$ is an explicit segment-level contribution to class $y$. The objective combines classification loss with a cosine concept-alignment loss between normalized learned concept activations and the frozen CLIP targets.
    }
    \label{fig:model_pipeline}
\end{figure*}

\textbf{Problem Setup:} Given a dataset $\mathcal{D} = \{(\myvec{x}_i, y_i)\}_{i=1}^{N_{\text{img}}}$ of $N_{\text{img}}$ images $\myvec{x}_i$ and class labels $y_i \in \mathcal{Y}$, standard classifiers learn a function from images to labels but provide little visibility into the concepts or regions driving each decision. Our goal is to design a model that represents images with semantically meaningful concepts, grounds those concepts in spatial regions, and aggregates region-level evidence through a computation that can be read back as an explanation. SEG-MIL-CBM has four stages, all defined below: an offline preprocessing pipeline turns each image into a bag of concept-guided segments and frozen concept targets; a trainable segment encoder and concept head produce learned concept activations for each segment; attention-based MIL aggregates segment-class evidence into image-level logits; and explanations are obtained by reading the positive segment contributions from the same logit decomposition.

\textbf{Preprocessing and Concept-Based Masking:} All segments consumed by the classifier are generated by a preprocessing pipeline and cached before model training. During training and evaluation on the cached benchmarks, the trainable model consumes only the cached segment bags and frozen CLIP concept-similarity targets; for a new uncached image, the same CLIP, GroundingDINO, and SAM preprocessing must be run once before downstream classification. Given an image $\myvec{x}$ and a concept vocabulary $\mathcal{C}=\{c_1,\dots,c_K\}$ of $K$ candidate concepts, we define the CLIP score of concept $c_k$ for any image crop $\myvec{u}$ as
\[
q_k(\myvec{u}) =
\frac{\exp(\cos(\psi_{\mathrm{img}}(\myvec{u}),\psi_{\mathrm{text}}(c_k))/\tau_{\mathrm{CLIP}})}
{\sum_{r=1}^{K}\exp(\cos(\psi_{\mathrm{img}}(\myvec{u}),\psi_{\mathrm{text}}(c_r))/\tau_{\mathrm{CLIP}})},
\]
where $\psi_{\mathrm{img}}$ and $\psi_{\mathrm{text}}$ are frozen CLIP image and text encoders and $\tau_{\mathrm{CLIP}}$ is the CLIP softmax temperature. The pipeline proceeds in three stages:
\begin{enumerate}[noitemsep,topsep=0pt,leftmargin=*]
    \item \textbf{Concept scoring.} We compute $q_k(\myvec{x})$ for all $c_k\in\mathcal{C}$ and keep the top-$K_{\text{top}}$ concepts $\mathcal{C}_{\myvec{x}}\subset\mathcal{C}$. Here $K_{\text{top}}$ is the number of concepts queried for that image. We follow the concept-list protocol of Label-Free-CBM~\cite{oikarinenlabel} for $\mathcal{C}$.
    \item \textbf{Grounded detection and segmentation.} For each $c\in\mathcal{C}_{\myvec{x}}$, GroundingDINO~\cite{liu2024grounding} proposes boxes localizing $c$ in $\myvec{x}$. Each box is segmented by SAM~\cite{kirillov2023segment} into a binary mask $\myvec{m}_i$ and converted into a masked segment image $\myvec{s}_i$. We then compute a frozen target vector $\myvec{c}^{\mathrm{CLIP}}_i=(q_1(\myvec{s}_i),\dots,q_K(\myvec{s}_i))\in\mathbb{R}^K$ over the full concept vocabulary.
    \item \textbf{Filtering and merging.} We discard masks that cover fewer than $\tauminpix$ pixels or more than a fraction $\rho_{\max}$ of the image, and merge masks whose pairwise IoU exceeds $\tauIoU$.
\end{enumerate}
The output for image $\myvec{x}$ is the \emph{bag}
\[
\mathcal{B}_{\myvec{x}}=\{(\myvec{s}_i,\myvec{m}_i,b_i,\myvec{c}_i^{\mathrm{CLIP}},r_i)\}_{i=1}^{N_s},
\]
where $\myvec{s}_i$ is the $i$th masked segment image, $\myvec{m}_i$ its binary mask, $b_i$ its bounding box, $\myvec{c}_i^{\mathrm{CLIP}}\in\mathbb{R}^K$ its frozen CLIP concept-similarity vector, $r_i=|\myvec{m}_i|/|\myvec{x}|$ its relative area, and $N_s$ the number of retained segments for $\myvec{x}$. The bag is cached to disk once per dataset and consumed as input by the MIL training loop. Full preprocessing diagrams, thresholds, pseudocode, and sensitivity analyses are in Appendix~\ref{appendix:technical}, Appendix~\ref{app:impl}, and Appendix~\ref{app:sensitivity}.

\textbf{Model Training and Accountable Aggregation:} After decomposing each image into a set of concept-guided segments, we treat the image as a bag of instances and train the model under a Multiple Instance Learning (MIL) framework in the style of attention-based MIL~\cite{ilse2018attention}. The trainable model has three components: a shared feature extractor $\phi$ that encodes each segment, a linear concept head that projects segment features into concept space, and an attention module that assigns weights to segments and aggregates them into an image-level prediction. Using the cached bag, the segment feature is $\myvec{h}_i=\phi(\myvec{s}_i)\in\mathbb{R}^d$, and the learned concept activation is $\myvec{z}_i=W_c\myvec{h}_i+\myvec{b}_c\in\mathbb{R}^K$. Here $\myvec{z}_i$ is used for prediction, while $\myvec{c}_i^{\mathrm{CLIP}}$ is used only as a frozen semantic alignment target. The attention network produces a class-conditioned score
\[
\begin{aligned}
u_{i,y}
&=\frac{\myvec{a}_y^\top \tanh(A\myvec{h}_i+\myvec{d})+e_y}
        {\tau_{\mathrm{attn}}},\\
\alpha_{i,y}
&=\frac{\exp(u_{i,y})}{\sum_{j=1}^{N_s}\exp(u_{j,y})},
\end{aligned}
\]
where $A,\myvec{d},\myvec{a}_y,e_y$ are learned attention parameters and $\tau_{\mathrm{attn}}$ is the attention temperature. The normalized attention weights satisfy $\sum_{i=1}^{N_s}\alpha_{i,y}=1$ for every class $y$. The class-conditioned configuration uses separate attention scores for each class. The shared-attention variant drops the class index and sets $\alpha_{i,y}=\alpha_i$ for all $y$; the mean-pooling ablation sets $\alpha_{i,y}=1/N_s$.

A linear classifier with row $\myvec{w}_y$ and bias $b_y$ maps each learned segment concept vector to a raw segment-class score $\bar g_{i,y}=\myvec{w}_y^\top\myvec{z}_i+b_y$. When area normalization is enabled, this score is rescaled by
\[
\eta_i=
\frac{\max(r_i,r_{\min})^{-\gamma}}
{\frac{1}{N_s}\sum_{j=1}^{N_s}\max(r_j,r_{\min})^{-\gamma}},
\qquad
g_{i,y}=\eta_i\bar g_{i,y},
\]
where $r_{\min}$ prevents tiny masks from dominating and $\gamma$ controls normalization strength; otherwise $\eta_i=1$. The image-level logit is $\ell_y = \sum_i \alpha_{i,y}\,g_{i,y}$ and $\hat y=\arg\max_y \ell_y$. The classifier bias is a class-level offset rather than visual evidence; for spatial explanations we therefore separate the exact logit contribution from the non-bias regional evidence described next.

To preserve semantic meaning, we align normalized learned concept activations with normalized frozen CLIP targets using the cosine loss $\mathcal{L}_{\mathrm{concept}} = -M^{-1}\sum_i \cos(\tilde{\myvec{z}}_i, \tilde{\myvec{c}}_i^{\mathrm{CLIP}})$, where \(M\) is the number of retained segments in the mini-batch, $\tilde{\myvec{z}}_i=\myvec{z}_i/\|\myvec{z}_i\|_2$, and $\tilde{\myvec{c}}_i^{\mathrm{CLIP}}=\myvec{c}_i^{\mathrm{CLIP}}/\|\myvec{c}_i^{\mathrm{CLIP}}\|_2$. The classification term $\mathcal{L}_{\mathrm{cls}}$ is image-level cross-entropy on the logits $\{\ell_y\}_{y\in\mathcal{Y}}$, and the model is trained with $\mathcal{L}_{\mathrm{total}} = \mathcal{L}_{\mathrm{cls}} + \lambda_{\mathrm{concept}}\,\mathcal{L}_{\mathrm{concept}}$, where $\lambda_{\mathrm{concept}}$ controls the strength of concept alignment. Group labels are not used in this objective.

\textbf{Spatial Importance:} For prediction-conditioned explanations, let \(C^{\mathrm{logit}}_{i,y}=\alpha_{i,y}g_{i,y}\) denote the exact segment term used by the forward pass. Since $b_y$ contains no localized evidence, displayed spatial support and concept names use the non-bias evidence \(C^{\mathrm{sp}}_{i,y}=\alpha_{i,y}\eta_i\,\myvec{w}_y^\top\myvec{z}_i\), with the bias treated as a global offset. For predicted class $\hat y$, the displayed segment importance is $S_i^{\mathrm{imp}}=\max(0,C^{\mathrm{sp}}_{i,\hat y})$; if all spatial contributions are non-positive, we rank by absolute spatial contribution. To name concept evidence inside a segment, we decompose $C^{\mathrm{sp}}$ into $E_{i,k}^{(\hat y)}=\alpha_{i,\hat y}\,\eta_i\,w_{\hat y,k}\,z_{i,k}$, whose positive values identify concepts supporting the prediction.

To obtain a spatial map, we distribute segment importance scores across the pixel supports of their corresponding masks as $H_{\mathrm{imp}}(p) \propto \sum_i S_i^{\mathrm{imp}}\; \mathbf{1}\{p \in \mathrm{mask}_i\}$, where \(p\) indexes image pixels, $\mathrm{mask}_i$ is the pixel support of $\myvec{m}_i$, and \(\mathbf{1}\{\cdot\}\) is an indicator function, followed by normalization for visualization. This produces a heatmap that highlights the regions that most strongly support the predicted class (see SEG-MIL-CBM heatmaps in Figure \ref{fig:opening_figure}).

\section{Experiments}
\label{sec:experiments}

We evaluate SEG-MIL-CBM with a primary focus on faithful, spatially grounded concept reasoning. The central question is whether the model's region-level accounting is both interpretable and tied to the prediction pathway. In addition to interpretability, we use spurious-correlation benchmarks to evaluate whether region-level aggregation improves worst-group behavior while preserving transparent concept-level explanations.

\subsection{Datasets}
We evaluate SEG-MIL-CBM on three benchmark families: spurious-correlation robustness, standard recognition, and corruption robustness. For spurious-correlation robustness, we use \textbf{Waterbirds}~\cite{sagawa2019distributionally}, which combines CUB bird images~\cite{wah2011caltech} with Places scene backgrounds~\cite{zhou2017places}. This construction induces background--label dependencies (water vs.\ land) that encourage shortcut learning. The Waterbirds test split includes bird-type $\times$ background groups, including mismatched groups, enabling a worst-group analysis. For Waterbirds and Pawrious, group labels are not used in the training objective; they are used only to compute validation and test metrics.

\paragraph{Pawrious.}
\textbf{Pawrious} is a synthetic spurious-correlation benchmark we construct from a customized variant of the Stable-Diffusion-based Spawrious generator~\cite{lynch2023spawrious}. Each image depicts a generated dog rendered against one of two natural backgrounds: jungle or snow. The target label is binary, grouping dogs into \emph{companion} and \emph{working} classes. The training distribution introduces a strong, asymmetric class--background coupling: companion dogs appear on snow in $90\%$ of training examples and jungle in $10\%$, while working dogs appear on jungle in $90\%$ of training examples and snow in $10\%$. The held-out splits contain all four class--background groups, exposing models that exploit background as a shortcut. A sample grid illustrating the bias structure is shown in Appendix~\ref{appendix:pawrious}.

To assess clean recognition, we include \textbf{CIFAR-10}~\cite{krizhevsky2009learning}, \textbf{CIFAR-100}~\cite{krizhevsky2009learning}, \textbf{CUB-200-2011 (CUB)}~\cite{wah2011caltech}, \textbf{Places365}~\cite{zhou2017places}, and \textbf{ImageNet (ILSVRC 2012)}~\cite{deng2009imagenet}. Robustness to common corruptions is measured with \textbf{CIFAR-10-C}~\cite{hendrycks2019robustness}, which aggregates 15 corruption types across five severities; we follow the standard mCE protocol and additionally report severity-conditioned accuracy in the appendix.

\subsection{Baselines}
To contextualize SEG-MIL-CBM, we compare against representative methods grouped as in Table~\ref{tab:comparison_criteria}: concept-based models without spatial localization, spatially aware CBMs, and representative non-CBM group-robust training methods.

\textbf{CBMs without spatial localization:} Post-hoc CBM~\cite{yuksekgonul2023posthoc} retrofits concept predictions after training; Label-Free-CBM~\cite{oikarinenlabel} discovers concepts without manual annotations; LaBo~\cite{yang2023language} uses language-defined bottlenecks; CDM~\cite{panousis2023sparse} learns sparse, disentangled concepts; DCLIP~\cite{menon2022visual} adapts CLIP features into a bottleneck; and DN-CBM~\cite{rao2024discover} discovers novel concepts dynamically. These approaches expose which concepts are used but not where they are supported in the image.

\textbf{Spatially aware CBMs and robust methods:} SALF-CBM~\cite{benou2025show} produces concept maps that localize evidence, and DCBM~\cite{dcbm2025} builds predictions over discovered regions/proposals. These models offer spatial grounding but limited machinery for quantifying how individual regions contribute to the final prediction. We also include GroupDRO~\cite{sagawa2019distributionally} and DFR~\cite{kirichenko2022last} in the main spurious-correlation table as non-interpretable robustness references. Extended group-robust baselines appear in Appendix~\ref{appendix:group_robust_baselines}.

\subsection{Experimental Setup}
For Table~\ref{tab:spurious_eval_cbm} (Waterbirds, Pawrious), we train SEG-MIL-CBM using a ResNet-50 image backbone, class-conditioned segment attention, and CLIP ViT-B/32 as the CLIP image backbone. For Waterbirds, we report a fixed configuration averaged over five seeds; for Pawrious, we use three independent seeds. For Table~\ref{tab:clip_vitb16}, to align with SALF-CBM~\cite{benou2025show} and DCBM~\cite{dcbm2025}, we swap the image backbone to ViT-B/16. All other hyperparameters are identical to those used for Table~\ref{tab:spurious_eval_cbm}. For prior spatial CBMs, we use the most complete reported numbers available from the corresponding papers and note backbone differences in the table caption.

For the Waterbirds component ablation in Table~\ref{tab:waterbirds_component_ablation}, we use the ViT-B/16 Waterbirds setup and run three seeds ($7200$--$7202$). The full model uses concept-guided segments, class-conditional learned attention, and $\lconcept=0.1$. We compare against computing attention from learned concept activations $\myvec{z}_i$ rather than unconstrained segment embeddings $\myvec{h}_i$, removing the concept-alignment loss, replacing learned attention with uniform mean pooling over segments, and replacing the segment bag with a single full-image instance while keeping the same backbone and concept head.

Unless otherwise noted, we report results directly from the respective papers. We re-implemented and ran the official code for Label-Free-CBM~\cite{oikarinenlabel} and Post-hoc-CBM~\cite{yuksekgonul2023posthoc} (in Table~\ref{tab:spurious_eval_cbm} and CIFAR-10-C), as well as AFR~\cite{qiu2023simple}, JTT~\cite{liu2021just}, and GroupDRO~\cite{sagawa2019distributionally} for the spurious-correlation experiments. Because the public SALF-CBM release did not expose full training code at the time of our experiment, we implemented SALF-CBM locally using the closest matching ViT-B/16 Waterbirds protocol.

\subsection{Evaluation Metrics}
\textbf{Interpretability and spatial grounding:} Since our primary goal is explainable concept-based prediction, we evaluate whether the model's region rankings faithfully reflect its own decision process. We use segment-level deletion and insertion on CUB, removing or inserting regions according to their predicted importance and reporting AUC-based faithfulness metrics. Qualitative examples illustrate whether concept evidence is spatially grounded in meaningful image regions.

\textbf{Predictive accuracy and robustness:} Standard average accuracy measures overall classification performance across all test samples. To evaluate robustness to spurious correlations, we consider latent subgroups $\mathcal{G}$, such as bird type $\times$ background in Waterbirds, and report worst-group accuracy, $\min_{g \in \mathcal{G}} \mathbb{E}_{(\myvec{x}, y) \sim \mathcal{D}_g}[\mathbb{1}\{f_\theta(\myvec{x}) = y\}]$. For CIFAR-10-C, we report mCE and severity-conditioned accuracy in Appendix~\ref{appendix:CIFAR10-C_results}.

\section{Results}

\begin{table}[t]
\centering
\scriptsize
\setlength{\tabcolsep}{0pt}
\begin{tabular*}{\columnwidth}{@{\extracolsep{\fill}}lccccc@{}}
\toprule
 & IMN & Places & CUB & CIFAR-10 & CIFAR-100 \\
\midrule
Linear Probe & 80.2 & 55.1 & 81.0 & 96.2 & 86.4 \\
Zero Shot & 68.6 & 39.5 & 55.0 & 91.6 & 68.7 \\
\midrule
Label-Free-CBM & 75.4 & 48.2 & 74.0 & 94.7 & 77.4 \\
Post-hoc-CBM & -- & -- & 61.0 & 87.1 & 68.0 \\
LaBo & 78.9 & -- & -- & \underline{95.7} & 81.2 \\
CDM & 79.3 & \underline{52.6} & \textbf{79.5} & 95.3 & \underline{80.5} \\
DCLIP & 68.0 & 40.3 & 57.8 & -- & -- \\
DN-CBM & \textbf{79.5} & \textbf{55.1} & -- & \textbf{96.0} & 82.1 \\
\midrule
DCBM-SAM2 & 70.4 & 50.6 & 75.3 & 95.2 & 79.4 \\
DCBM-GDINO & 69.7 & 50.7 & 74.1 & 95.1 & 79.6 \\
DCBM-MASKRCNN & 70.5 & 50.9 & 76.7 & 95.2 & 79.6 \\
SALF-CBM & \underline{78.6} & 49.4 & 76.2 & -- & -- \\
\rowcolor[HTML]{C6EAD8} SEG-MIL-CBM (ours) & 78.5$\pm$0.2 & 51.24$\pm$0.25 & \underline{77.39$\pm$0.22} & 94.89$\pm$0.12 & \textbf{85.26$\pm$0.00} \\
\bottomrule
\end{tabular*}
\caption{
    Accuracy with CLIP ViT-B/16 backbone. Baselines are from DCBM~\cite{dcbm2025} unless noted; method citations are given in Section~\ref{sec:experiments}, and SALF-CBM~\cite{benou2025show} uses the reported backbone for each dataset. Mean~$\pm$~std is shown for SEG-MIL-CBM over independent runs; other entries are single reported values. \textbf{Bold} marks best and \underline{underline} second best.
}
\label{tab:clip_vitb16}
\end{table}

\begin{table}[t]
\centering
\footnotesize
\setlength{\tabcolsep}{1pt}
\begin{tabular*}{\columnwidth}{@{\extracolsep{\fill}}lcccc@{}}
\toprule
\textbf{Model} & \multicolumn{2}{c}{\textbf{Waterbirds (\%)}} & \multicolumn{2}{c}{\textbf{Pawrious (\%)}} \\
 & Avg. & Worst & Avg. & Worst \\
\midrule
ERM & \textbf{97.3} & 60.0 & \textbf{98.59} & 75.55 \\
\midrule
\multicolumn{5}{@{}l}{\emph{Non-CBM robust methods}} \\
GroupDRO \cite{sagawa2019distributionally} & 96.0 & 86.0 & 90.83 & 86.67 \\
DFR \cite{kirichenko2022last} & 94.2 & \textbf{92.9} & -- & -- \\
\midrule
\multicolumn{5}{@{}l}{\emph{Global concept baselines}} \\
Label-Free-CBM \cite{oikarinenlabel} & 81.82 & 54.62 & 94.67 & 46.67 \\
Post-hoc-CBM \cite{yuksekgonul2023posthoc} & 80.58 & 57.89 & 91.20 & 19.26 \\
\midrule
\multicolumn{5}{@{}l}{\emph{Spatially grounded CBMs}} \\
SALF-CBM$^\dagger$ \cite{benou2025show} & 78.89 & 65.11 & -- & -- \\
\rowcolor[HTML]{C6EAD8} SEG-MIL-CBM (ours) & 85.73$\pm$0.18 & 72.00$\pm$1.08 & 97.73$\pm$0.01 & \textbf{87.41$\pm$0.20} \\
\bottomrule
\end{tabular*}
\caption{
    Spurious-correlation results on Waterbirds and Pawrious. Avg and Worst denote average and worst-group accuracy. Mean~$\pm$~std is shown for SEG-MIL-CBM over independent runs; other entries are single reported or locally reproduced values. $^\dagger$ We implemented SALF-CBM locally because full training code was unavailable. Full group-robust results are in Appendix~\ref{appendix:group_robust_baselines}, Table~\ref{apptab:spurious_eval_dnn}.
}
\label{tab:spurious_eval_cbm}
\end{table}

\begin{table}[t]
\centering
\footnotesize
\setlength{\tabcolsep}{1pt}
\begin{tabular*}{\columnwidth}{@{\extracolsep{\fill}}lcc@{}}
\toprule
\textbf{Model} & \textbf{Avg. (\%)} & \textbf{Worst-group (\%)} \\
\midrule
\rowcolor[HTML]{C6EAD8} Full SEG-MIL-CBM & 85.38$\pm$0.15 & 71.29$\pm$1.26 \\
Concept-attention SEG-MIL-CBM & 85.35$\pm$0.17 & 72.17$\pm$0.89 \\
No concept alignment ($\lconcept=0$) & 85.26$\pm$0.27 & 71.98$\pm$0.59 \\
Mean-pool segments & 82.05$\pm$0.20 & 66.50$\pm$0.40 \\
Global no-segment CBM & 85.05$\pm$0.17 & 68.69$\pm$1.12 \\
\bottomrule
\end{tabular*}
\caption{
    \textbf{Waterbirds component ablation.} Mean~$\pm$~std over three ViT-B/16 runs. Variants compute attention from concepts, remove concept alignment, replace attention with mean pooling, or use a full-image no-segment instance.
}
\label{tab:waterbirds_component_ablation}
\end{table}

\begin{table}[t]
\centering
\footnotesize
\setlength{\tabcolsep}{1pt}
\begin{tabular*}{\columnwidth}{@{\extracolsep{\fill}}lcccc@{}}
\toprule
\textbf{Ranking} & \textbf{DAUC} $\downarrow$ & \textbf{IAUC} $\uparrow$ & \textbf{Top-1 Ins.} $\uparrow$ & \textbf{Top-5 Del.} $\uparrow$ \\
\midrule
Raw CLIP segment score & 0.486 & 0.656 & 84.9 & 17.6 \\
Classifier contribution only & 0.474 & 0.650 & 83.5 & 19.8 \\
Segment area & 0.475 & 0.657 & 84.8 & 18.3 \\
Random & 0.555 & 0.620 & 68.2 & 6.9 \\
Same-area random & 0.497 & 0.647 & 83.3 & 15.7 \\
\rowcolor[HTML]{C6EAD8} SEG-MIL-CBM importance & \textbf{0.444} & \textbf{0.670} & \textbf{90.5} & \textbf{22.2} \\
\bottomrule
\end{tabular*}
\caption{
    \textbf{Segment-level faithfulness evaluation on CUB.} Deletion/insertion is measured on $3000$ test images after zeroing selected segment embeddings. DAUC is lower better; IAUC, Top-1 insertion, and Top-5 deletion are higher better.
}
\label{tab:interp_eval}
\end{table}

\begin{figure*}[t]
    \centering
    \includegraphics[width=0.85\textwidth]{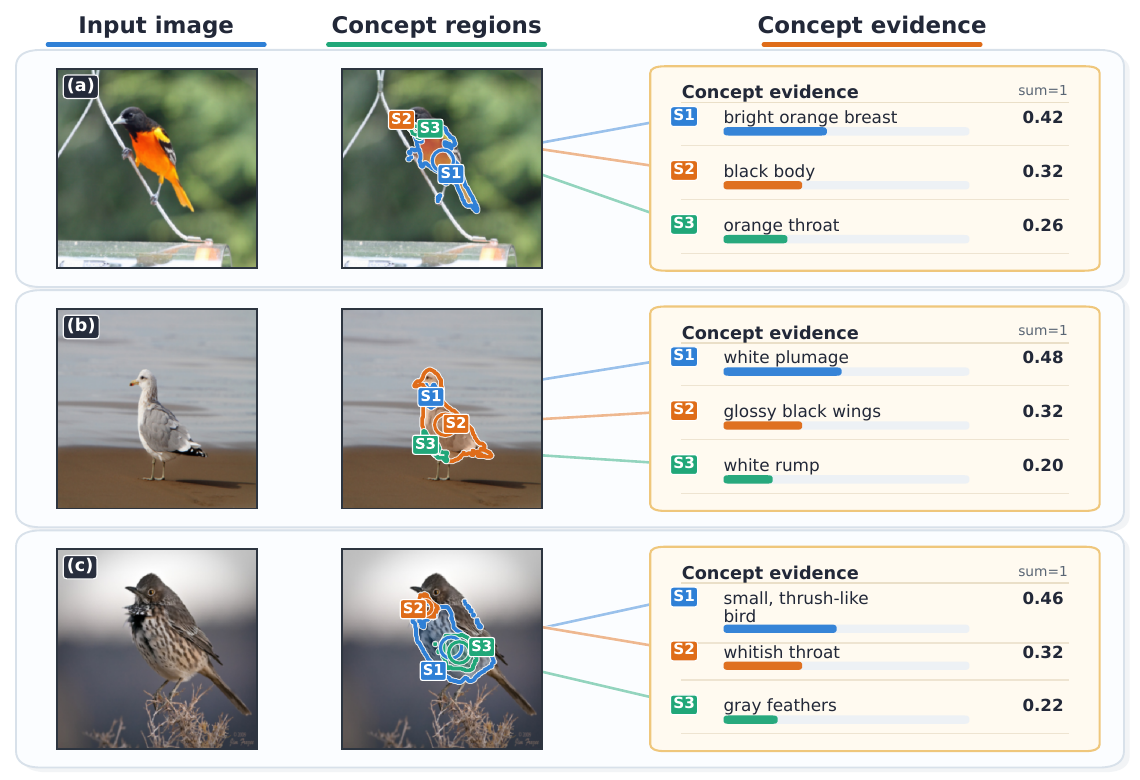}
    \caption{
        \textbf{Qualitative segment-level explanations.} Each row shows an input CUB image, the concept-guided regions selected by SEG-MIL-CBM, and the corresponding concept evidence used for the image-level prediction. For each highlighted segment, the evidence panel reports the strongest positive concept activation and its normalized segment contribution; displayed contributions sum to one within each example.
    }
    \label{fig:interp_qualitative_grid}
\end{figure*}

We evaluate SEG-MIL-CBM along three axes: segment-level faithfulness (Table~\ref{tab:interp_eval}, Figure~\ref{fig:interp_qualitative_grid}), standard recognition accuracy (Table~\ref{tab:clip_vitb16}), and group robustness under spurious correlations (Table~\ref{tab:spurious_eval_cbm}). We also include a Waterbirds component ablation (Table~\ref{tab:waterbirds_component_ablation}) to isolate the contribution of segments, learned attention, and concept alignment.

\textbf{Segment-level faithfulness.} Table~\ref{tab:interp_eval} is our primary faithfulness test. Following deletion/insertion protocols, we rank segments by decreasing $S_i^{\mathrm{imp}}$ and measure how the model behaves when evidence is removed or reintroduced. For deletion, we progressively set the top-ranked segment embeddings to zero and rerun the model on the perturbed bag. For insertion, we start from an all-zero segment bag, restore only the top-ranked segment embeddings, and rerun the model. In both cases, the attention weights, segment-class scores, classifier bias terms, and final logits are recomputed from the perturbed bag rather than held fixed from the original forward pass. This avoids pixel-grid artifacts and evaluates faithfulness at the native explanation unit of SEG-MIL-CBM.

SEG-MIL-CBM achieves the best DAUC/IAUC (0.444/0.670), improving over raw CLIP segment salience, classifier-only contribution, segment area, random ranking, and same-area random controls. The top-$k$ interventions tell the same story: keeping only the highest-ranked SEG-MIL-CBM segment preserves the original prediction in $90.5\%$ of images, compared with $84.9\%$ for raw CLIP and $68.2\%$ for random; zeroing the top five SEG-MIL-CBM segments flips $22.2\%$ of predictions, compared with $17.6\%$ for raw CLIP and $6.9\%$ for random. An additional attention-only ranking control is reported in Appendix~\ref{appendix:attention_only_faithfulness}. It gives nearly identical deletion/insertion curves, indicating that this perturbation protocol primarily measures faithful region selection. We therefore use the CUB attribute-agreement diagnostic in Appendix~\ref{appendix:cub_attribute_agreement} to evaluate the semantic contribution of the concept-conditioned term.

\textbf{Localization and concept semantics.} CUB-200-2011 provides part annotations for visible bird anatomy. We use these annotations as a saturation check for object-centric localization rather than as the main faithfulness test. SEG-MIL-CBM localizes far above a random-pixel baseline ($56.0\%$ vs.\ $4.4\%$) and matches raw CLIP segment salience ($55.9\%$), indicating that SAM and CLIP already produce strong object-centric candidate segments on CUB. Segment-level faithfulness is therefore the more discriminative test of whether the learned ranking matters for the model's prediction. To test whether the displayed concepts are semantically meaningful, Appendix~\ref{appendix:cub_attribute_agreement} maps the fixed free-form concept vocabulary to CUB attributes only for evaluation. SEG-MIL-CBM concept contributions improve attribute-ranking mAP over an attention-attached-name control from $17.1$ to $18.4$, and over random attributes ($11.2$), supporting the role of the concept-conditioned classifier term.

\textbf{Global sparsity diagnostic.} We also probe whether the full CUB classifier can be compressed to a tiny global concept set by adding an $\ell_1$ penalty on $W_y$ and sweeping the sparsity level, following Srivastava~\etal~\cite{srivastava2024vlgcbm}. This transparency diagnostic shows a clear trade-off: constraining the model to approximately five active concepts drives performance close to chance, while weaker sparsity settings still use most concepts. This is consistent with fine-grained 200-way bird recognition, where many species share spatial support but require different subtle cues. Our interpretability claim is therefore local segment-level faithfulness and per-example concept accounting, rather than compression of the entire task into a small universal concept list.

\textbf{Standard recognition tasks.} Table~\ref{tab:clip_vitb16} shows that SEG-MIL-CBM delivers strong performance across benchmarks, leading among spatially aware CBMs and achieving the best CBM result on CIFAR-100. It also improves over prior spatial CBMs on Places365 and CUB. On ImageNet and CIFAR-10, it remains within a narrow margin of the top results. These results indicate that aggregating concept-aligned segments preserves large-scale recognition while exposing faithful segment-level evidence, especially for categories in which localized parts, such as beak, crown, or wing patterns, are discriminative.

\textbf{Group robustness under spurious correlations.} Table~\ref{tab:spurious_eval_cbm} shows the accuracy--interpretability trade-off directly. On Waterbirds, group-robust methods such as DFR and GroupDRO achieve higher worst-group accuracy ($92.9\%$ and $86.0\%$), but they do not provide concept-level or spatial explanations. Among evaluated CBM-family baselines, SEG-MIL-CBM improves worst-group accuracy from $65.11\%$ for the strongest baseline, SALF-CBM$^\dagger$, to $72.00\%$, making it the strongest method in this table that also provides spatially grounded concept accounting. SALF-CBM$^\dagger$ with the closest ViT-B/16 protocol reaches $78.89\%$ average accuracy, suggesting that spatial concept maps alone do not guarantee shortcut robustness when region-level aggregation is not explicitly accounted for. On Pawrious, the global CBM baselines fall below ERM in worst-group accuracy, consistent with their image-level concept vectors preserving the highly predictive snow/jungle background shortcut. By scoring localized concept evidence, SEG-MIL-CBM reaches $87.41\%$ worst-group accuracy while preserving high average accuracy.

\textbf{Component ablation and robustness diagnostics.} Table~\ref{tab:waterbirds_component_ablation} supports the role of segment aggregation. Computing attention from learned concept activations rather than unconstrained segment embeddings preserves performance ($85.35\%$ avg., $72.17\%$ worst), indicating that the Waterbirds gain does not require a hidden-feature attention bypass. Mean pooling and a single full-image instance both reduce worst-group accuracy, while removing concept alignment stays within run-to-run variability; we therefore interpret $\lconcept$ primarily as preserving the semantic concept interface used for explanation. As a background-reliance diagnostic on $1000$ Waterbirds test images, the foreground-union variant preserves performance and improves worst-group accuracy relative to the full cached bag ($84.1/78.1$ vs.\ $84.7/70.5$ avg./worst), while the background-complement variant drops sharply ($61.3/30.3$). Full protocols are in Appendix~\ref{appendix:concept_attention} and Appendix~\ref{appendix:foreground_background}; CIFAR-10-C corruption results are in Appendix~\ref{appendix:CIFAR10-C_results}.

\section{Discussion}
\label{sec:discussion}

\textbf{What the explanation does and does not certify.} SEG-MIL-CBM explains a prediction by decomposing the predicted logit into segment-level terms. This is stronger than attaching a heatmap to a separately trained classifier, because displayed region evidence is read from the same forward computation, with the non-spatial classifier bias kept separate. It also clarifies the scope of the claim: deletion/insertion tests whether the ranking is faithful to the model's own prediction behavior, not whether the model has recovered a complete human causal model of the scene. The explanation is therefore best read as an accountable trace of model evidence: which regions were routed to the predicted class, which concepts supplied positive support, and how much each region contributed relative to the other retained segments.

\textbf{Why segment aggregation matters.} The Waterbirds and Pawrious results suggest that explicit region aggregation can help when the shortcut is spatially separable from the object evidence. A global CBM can still encode a snow, water, or jungle concept in the same image-level vector that also represents the object; this makes it difficult to see or control whether the final classifier is relying on the shortcut. SEG-MIL-CBM instead scores localized regions and aggregates them through class-conditioned attention. When the relevant object evidence and the spurious background evidence occupy different segments, the model can route class evidence away from shortcut regions and expose that choice in the explanation. This does not guarantee shortcut removal in all cases: if the segmenter misses the object, merges object and background, or produces concepts that conflate them, the downstream model inherits that failure. The contribution accounting nevertheless makes such failures more inspectable than in a global bottleneck.

\textbf{Region selection versus concept semantics.} Deletion/insertion evaluates whether selected segments affect the model prediction, while the reported concept contributions additionally attach class-conditioned semantic evidence to those segments. The attention-only appendix control confirms that these two questions can separate: attention is already strong for region selection, whereas the CUB attribute-agreement diagnostic probes whether the displayed concept evidence is semantically meaningful. This distinction is important because the intended explanation is not only a region ranking, but a per-region account of which concepts supplied positive class evidence.

\textbf{Accuracy--interpretability trade-off.} Non-CBM group-robust methods remain stronger on Waterbirds worst-group accuracy, especially when group labels or carefully tuned post-hoc reweighting are available. SEG-MIL-CBM is not meant to replace those methods when accuracy alone is the only objective. Its contribution is a different Pareto point: it improves over evaluated CBM-family baselines while preserving a spatially grounded concept account of each prediction. This distinction matters in settings where a practitioner needs to inspect whether the model relied on object parts, background, or other localized evidence, rather than only observe that the worst-group number improved.

\section{Conclusion}
We introduced SEG-MIL-CBM, a spatially grounded concept model in which prediction and explanation share the same region-level computation. By aligning segment regions with concepts and aggregating them through attention, the model exposes the positive concept evidence that contributes to its predicted logit. Segment-level faithfulness experiments show that this learned segment ranking is faithful to the model's prediction behavior, while the broader results show competitive large-scale recognition, improved worst-group accuracy over evaluated CBM baselines on spurious-correlation benchmarks, and CIFAR-10-C diagnostics suggesting improved robustness under several corruptions. Compared with non-CBM group-robust methods, SEG-MIL-CBM should be read as a Pareto point: it trades some worst-group accuracy on Waterbirds for spatially grounded concept accounting, while remaining especially strong on Pawrious.

\textbf{Limitations and future work.} SEG-MIL-CBM depends on CLIP, GroundingDINO, and SAM preprocessing. This cost is amortized offline but can limit real-time use, and failures in region proposal or concept coverage bound both accuracy and explanation completeness. Part of the localization quality therefore comes from pretrained ROI discovery rather than the MIL aggregator alone.

\textbf{Acknowledgments.} OL and RE were supported by the MOST grant No. 0007341.

\clearpage

{
    \small
    \bibliographystyle{ieeenat_fullname}
    \bibliography{main}
}

\clearpage
\appendix
\section{Technical Details of SEG-MIL-CBM}
\label{appendix:technical}

This appendix provides the full technical details of the SEG-MIL-CBM framework, including the model architecture, loss functions, and training algorithm.

\subsection{Model Architecture}  
Each input image $\myvec{x}$ is decomposed into a bag $\mathcal{B} = \{\myvec{s}_1, \dots, \myvec{s}_{N_s}\}$ of $N_s$ segment-level instances $\myvec{s}_i$, each corresponding to a concept-guided mask generated during preprocessing. For simplicity, we write $\mathcal{B}$ instead of $\mathcal{B}_{\myvec{x}}$ here. The model consists of the following components:
  
\begin{itemize}[noitemsep,topsep=0pt,leftmargin=*]
    \item \textbf{Feature Extractor $\phi$}: A pretrained backbone that maps each segment instance $\myvec{s}_i$ to a feature vector $\myvec{h}_i = \phi(\myvec{s}_i) \in \mathbb{R}^d$.
    \item \textbf{Concept Head}: A linear projection $\myvec{W}_c \in \mathbb{R}^{K \times d}$ maps each feature $\myvec{h}_i$ into a $K$-dimensional concept space:
    \[
    \myvec{z}_i = \myvec{W}_c \myvec{h}_i + \myvec{b}_c \in \mathbb{R}^K,
    \]
    where $\myvec{z}_i$ is the predicted concept activation vector for instance $\myvec{s}_i$.
    \item \textbf{Attention Module}: Learns importance weights over instances via a temperature-scaled softmax. We write the general class-indexed form as
    \[
    \alpha_{i,y} = \frac{\exp(a_y(\myvec{h}_i) / T)}{\sum_{j=1}^{N_s} \exp(a_y(\myvec{h}_j) / T)},
    \]
    where $a_y(\cdot)$ is the output for class $y$ of a one-hidden-layer attention MLP and $T$ is a temperature hyperparameter. The shared-attention variant is the special case $a_y(\cdot)=a(\cdot)$ for all classes.
    \item \textbf{Classifier}: Maps each segment concept vector to a segment-class score:
    \[
    g_{i,y}=\myvec{w}_y^\top \myvec{z}_i+b_y.
    \]
    Some robustness runs use an optional mask-area normalization factor $\eta_i$ on segment logits:
    \[
    \eta_i=\frac{\max(r_i,\rho_{\min})^{-\gamma_{\mathrm{area}}}}
    {\frac{1}{N_s}\sum_{j=1}^{N_s}\max(r_j,\rho_{\min})^{-\gamma_{\mathrm{area}}}},
    \]
    where $r_i$ is the mask area fraction; setting $\gamma_{\mathrm{area}}=0$ disables the factor and gives $\eta_i=1$. The image-level logit is
    \[
    \ell_y=\sum_{i=1}^{N_s}\alpha_{i,y}\,\eta_i\,g_{i,y}.
    \]
    For spatial explanations, we separate this exact logit contribution from the non-bias visual evidence, as detailed below.
\end{itemize}

\subsection{Bias Accounting in Spatial Explanations}
\label{appendix:bias_accounting}

The forward pass above gives an exact segment-logit contribution
\[
C^{\mathrm{logit}}_{i,y}
=\alpha_{i,y}\eta_i(\myvec{w}_y^\top\myvec{z}_i+b_y).
\]
However, $b_y$ is a class-level offset and does not correspond to localized visual evidence. We therefore use the non-bias term
\[
C^{\mathrm{sp}}_{i,y}
=\alpha_{i,y}\eta_i\,\myvec{w}_y^\top\myvec{z}_i
\]
for displayed spatial support and concept naming, and keep
\[
C^{\mathrm{bias}}_{i,y}
=\alpha_{i,y}\eta_i b_y
\]
as a global classifier-offset contribution. When area normalization is disabled, $\sum_i C^{\mathrm{bias}}_{i,y}=b_y$ because $\sum_i\alpha_{i,y}=1$; with area normalization, the bias term is still not a concept or region attribute. This convention prevents a segment from receiving semantic support solely by inheriting part of the classifier bias.

\subsection{Training Objective}  
The training objective combines two components:
\begin{enumerate}[noitemsep,leftmargin=*]
    \item \textbf{Classification Loss}: Standard cross-entropy loss for predicting the correct image-level label.
    \item \textbf{Concept Alignment Loss}: A cosine similarity loss between normalized predicted concept activations $\tilde{\myvec{z}}_i=\myvec{z}_i/\|\myvec{z}_i\|_2$ and normalized CLIP-derived similarity vectors $\tilde{\myvec{z}}_i^{\mathrm{CLIP}}=\myvec{z}_i^{\mathrm{CLIP}}/\|\myvec{z}_i^{\mathrm{CLIP}}\|_2$:
    \[
    \mathcal{L}_{\mathrm{concept}} = -\frac{1}{M} \sum_{i=1}^M \cos\left(\tilde{\myvec{z}}_i, \tilde{\myvec{z}}_i^{\mathrm{CLIP}}\right),
    \]
    where $M$ is the number of segment instances in the mini-batch.
\end{enumerate}
The total loss is defined as:
\[
\mathcal{L}_{\mathrm{total}} = \mathcal{L}_{\mathrm{cls}} + \lconcept \cdot \mathcal{L}_{\mathrm{concept}},
\]
where $\lconcept$ balances classification and concept alignment.

\subsection{Preprocessing and Training Algorithms}

\begin{figure*}[t]
    \centering
    \includegraphics[width=\textwidth]{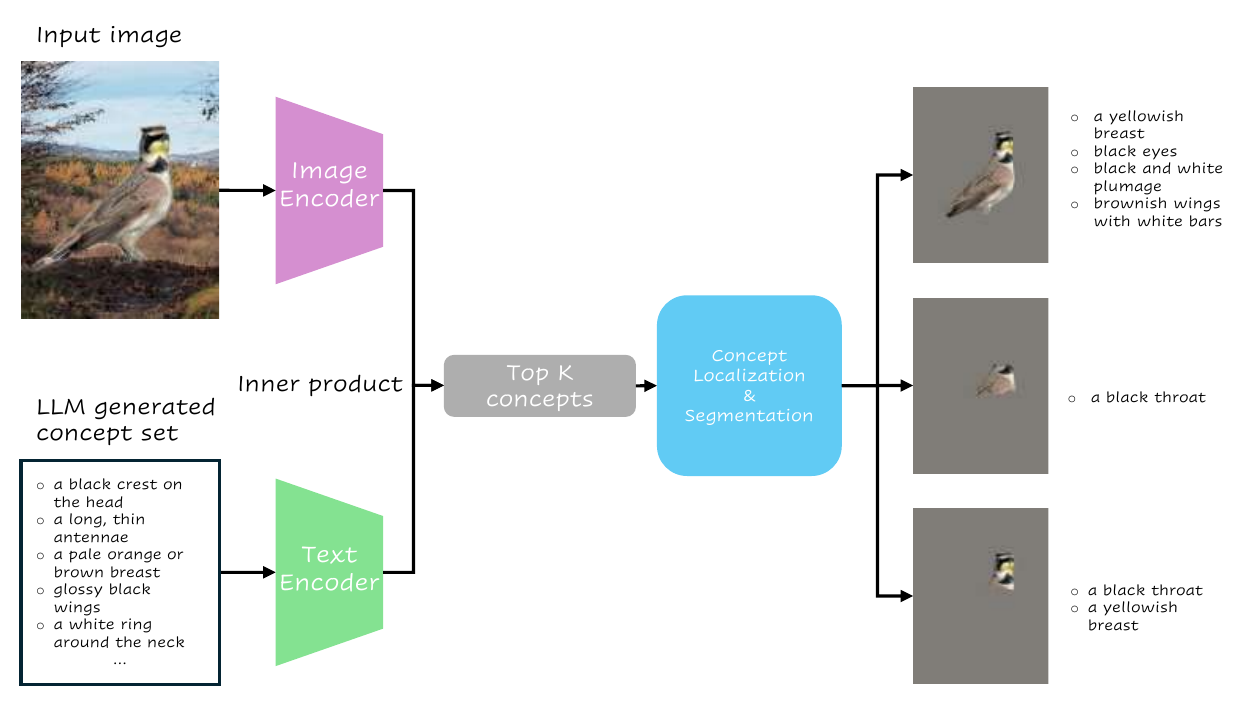}
    \caption{
        \textbf{Overview of our offline concept-guided segmentation pipeline.} Given an input image, a vision-language image encoder extracts image embeddings while the concept set is encoded by the corresponding text encoder. The top-$K_{\text{top}}$ concepts are selected using cosine-similarity scores and then passed to GroundingDINO and SAM to produce semantically meaningful segments. Each cached segment is annotated with concept phrases and its CLIP-derived similarity vector $\myvec{z}_i^{\mathrm{CLIP}}$.
    }
    \label{fig:concept_pipeline}
\end{figure*}

\begin{algorithm}[h]
\caption{
    CLIP \cite{radford2021learning} Guided Concept Segmentation and Bag Creation
}
\label{alg:clip_concept}
\begin{algorithmic}[1]
\REQUIRE Image dataset $\mathcal{D}$, concept list $\mathcal{C}$, pretrained models: CLIP \cite{radford2021learning}, GroundingDINO \cite{liu2024grounding}, SAM \cite{kirillov2023segment}.
\FOR{each image $\myvec{x} \in \mathcal{D}$}
    \STATE Compute CLIP \cite{radford2021learning} similarity scores: $\mathrm{CLIP}(\myvec{x}, c)$ for all $c \in \mathcal{C}$
    \STATE Select top-$K_{\text{top}}$ concepts $\mathcal{C}_{\myvec{x}} \subset \mathcal{C}$ based on similarity
    \STATE Initialize bag $\mathcal{B}_{\myvec{x}} = \{\}$
    \FOR{each concept $c \in \mathcal{C}_{\myvec{x}}$}
        \STATE Use GroundingDINO \cite{liu2024grounding} to detect bounding boxes for $c$ in $\myvec{x}$
        \FOR{each detected box $b$}
            \STATE Segment region inside $b$ using SAM  \cite{kirillov2023segment} $\rightarrow$ binary mask $\myvec{m}$
            \STATE Extract masked segment image $\myvec{s}$ and CLIP similarity vector $\myvec{z}^{\mathrm{CLIP}}$
            \STATE Add $(\myvec{s}, \myvec{m}, b, \myvec{z}^{\mathrm{CLIP}})$ to $\mathcal{B}_{\myvec{x}}$
        \ENDFOR
    \ENDFOR
    \STATE Merge overlapping masks in $\mathcal{B}_{\myvec{x}}$ (IoU $> \tauIoU$)
    \STATE Save $\mathcal{B}_{\myvec{x}}$
\ENDFOR
\RETURN $\{\mathcal{B}^{(j)}\}_{j=1}^{N_{\text{img}}}$
\end{algorithmic}
\end{algorithm}

\begin{algorithm}[h]
\caption{
    MIL Training with CLIP-Guided Concept Supervision (pedagogical version).
}
\label{alg:mil-training-appendix}
\begin{algorithmic}[1]
\REQUIRE Bags $\{(\mathcal{B}^{(j)}, y^{(j)}, \{\myvec{z}_i^{\mathrm{CLIP}}\})\}_{j=1}^{N_{\text{img}}}$,
         attention temperature $T$, concept-loss weight $\lconcept$,
         epochs $E$.
\FOR{epoch $e = 1, \dots, E$}
    \FOR{each bag $(\mathcal{B}, y, \{\myvec{z}_i^{\mathrm{CLIP}}\}) $}
        \STATE $\myvec{H} \leftarrow \phi(\mathcal{B})$
            \hfill\textit{// per-segment features via shared backbone}
        \STATE $\myvec{Z} \leftarrow \myvec{W}_c \myvec{H}$
            \hfill\textit{// linear projection to concept space}
        \STATE $\tilde{\myvec{z}}_i \leftarrow \myvec{Z}_i / \|\myvec{Z}_i\|_2$
            \hfill\textit{// L2-normalize for cosine alignment}
        \STATE $\tilde{\myvec{z}}_i^{\mathrm{CLIP}} \leftarrow \myvec{z}_i^{\mathrm{CLIP}} / \|\myvec{z}_i^{\mathrm{CLIP}}\|_2$
            \hfill\textit{// normalize cached CLIP scores}
        \STATE $\alpha_{i,y} \leftarrow \softmax_i(a_y(\myvec{H}_i) / T)$
            \hfill\textit{// attention over segments; shared variant uses one $a(\cdot)$}
        \STATE $g_{i,y} \leftarrow \eta_i\,(\myvec{w}_y^\top\myvec{Z}_i+b_y)$
            \hfill\textit{// segment-class scores, with optional area factor $\eta_i$}
        \STATE $\ell_y \leftarrow \sum_i \alpha_{i,y}\,g_{i,y}$
            \hfill\textit{// aggregate segment logits}
        \STATE $\hat{y} \leftarrow \softmax(\boldsymbol{\ell})$
            \hfill\textit{// image-level prediction}
        \STATE $\mathcal{L}_{\mathrm{cls}} \leftarrow \mathrm{CE}(\hat{y}, y)$
            \hfill\textit{// image-level classification loss}
        \STATE $\mathcal{L}_{\mathrm{concept}} \leftarrow
                -\tfrac{1}{|\mathcal{B}|}\sum_i \cos(\tilde{\myvec{z}}_i, \tilde{\myvec{z}}_i^{\mathrm{CLIP}})$
            \hfill\textit{// align segment concepts with CLIP supervision}
        \STATE $\mathcal{L} \leftarrow \mathcal{L}_{\mathrm{cls}} + \lconcept\, \mathcal{L}_{\mathrm{concept}}$
        \STATE update $\phi, \myvec{W}_c$, attention MLP, and classifier by SGD on $\mathcal{L}$
    \ENDFOR
\ENDFOR
\end{algorithmic}
\end{algorithm}

\paragraph{Training Heuristics: Easy/Hard Batch Alternation}

As an additional implementation detail, we experimented with alternating between batches of “easy” and “hard” samples during training. After a warm-up phase, “easy” batches consisted of high-confidence predictions that reinforced task-relevant concepts, while “hard” batches contained low-confidence or misclassified samples that encouraged the model to address underrepresented regions and spurious correlations.

\subsection{Backbone Warm-Up Prior to Embedding Extraction}
\label{appendix:warmup}

Before segment-level embedding extraction, we apply a lightweight warm-up of the vision backbone on the original training dataset. This consists of fine-tuning the backbone for a few epochs with a standard classification objective, using only the original labels and no additional annotations. After this stage, the backbone is frozen, and all subsequent training proceeds as described in the main method. This warm-up aims to stabilize feature representations for masked segments and mitigate mild distribution shifts between full and segmented images, without introducing new supervision or capacity.

\clearpage
\section{Pawrious Dataset Bias Structure}
\label{appendix:pawrious}

\begin{figure}[h]
    \centering
    \includegraphics[width=0.72\linewidth]{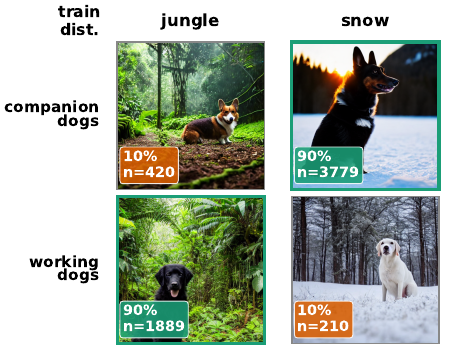}
    \caption{
        \textbf{Pawrious bias structure.} Rows denote the binary target classes and columns denote the spurious background attribute. Percentages and counts are computed on the training split. Majority shortcut pairs are highlighted in green; minority counterexamples remain in each class.
    }
    \label{fig:pawrious_grid}
\end{figure}

Figure~\ref{fig:pawrious_grid} summarizes the class--background coupling used to construct the Pawrious training split.

\clearpage
\section{Group-Robust Training Baselines}
\label{appendix:group_robust_baselines}

This appendix expands the non-CBM, group-robustness-focused comparison summarized in the main paper. The main text reports representative strong baselines to make the accuracy--interpretability trade-off explicit; here we provide the fuller set for completeness. SEG-MIL-CBM remains competitive in an interpretability-first setting and achieves the best worst-group accuracy on Pawrious.
On Pawrious, the strongest CBM baseline reaches $46.7\%$ worst-group accuracy, whereas SEG-MIL-CBM reaches $87.4\%$ while remaining within one percentage point of ERM average accuracy ($97.73\%$ vs.\ $98.59\%$), highlighting the robustness gain in the most shortcut-conflicting subgroup.

\paragraph{SALF-CBM Waterbirds run.}
Because, at the time of our experiment, the public SALF-CBM release did not expose full training code, we include a local SALF-style Waterbirds run using the closest matching protocol. The run uses a frozen ViT-B/16 image backbone, CLIP ViT-B/16 red-circle local visual prompts over a $7{\times}7$ grid, the CUB filtered concept vocabulary ($369$ concepts), a $1{\times}1$ spatial concept bottleneck trained with the cubic cosine objective, SALF softmax spatial pooling, and a sparse GLM-SAGA classifier. We select the classifier regularization by validation worst-group accuracy, with validation accuracy as a tie-breaker, matching the Waterbirds selection protocol used elsewhere in the paper. The selected setting uses $\lambda=10^{-5}$ and obtains $78.89\%$ average accuracy and $65.11\%$ worst-group accuracy on the Waterbirds test split. We mark this row with $^\dagger$ in the tables to distinguish it from numbers reported directly by the SALF-CBM paper.

\paragraph{Methods with group annotations.}
GroupDRO~\cite{sagawa2019distributionally} directly optimizes the worst-case group risk during training, improving reliability under spurious correlations but offering neither spatial localization nor concept-level interpretability. DFR~\cite{kirichenko2022last} mitigates shortcuts in a post-hoc stage by retraining only the final layer, again without concept or spatial explanations. DaC~\cite{Noohdani_2024_CVPR} pursues robustness through training-time decomposition and composition of features. Collectively, these methods are strong in mitigating spurious correlations when group labels are available, yet they provide limited transparency about what evidence is used.

\paragraph{Methods without group annotations.}
When explicit group labels are unavailable, several techniques infer structure or reweight data to improve worst-group behavior. DISC~\cite{wu23disc} leverages concept-aware counterfactual augmentation during training; CnC~\cite{zhang2022correct} encourages contrasting corrections; AFR~\cite{qiu2023simple} adjusts features in a post-hoc manner; EIIL~\cite{creager2021environment} infers pseudo-environments to enforce invariance; and JTT~\cite{liu2021just} reweights difficult examples via a second training pass. These methods target spurious-correlation robustness without requiring group supervision, but, like the group-annotated family, they do not yield concept-grounded or spatially localized explanations.

\begin{table}[h]
\scriptsize
\centering
\resizebox{\columnwidth}{!}{
\begin{tabular}{lccccc}
\toprule
\textbf{Method} & \textbf{Spatial} & \textbf{Stage} & \textbf{Concept} & \textbf{Spurious} & \textbf{Group-} \\
 & \textbf{Localization} & & \textbf{Interpretability} & \textbf{Mitigation} & \textbf{Annot.} \\
\midrule
GroupDRO \cite{sagawa2019distributionally} & \ding{55} & Training & \ding{55} & \ding{51} & \ding{51} \\
DFR \cite{kirichenko2022last} & \ding{55} & Post-hoc & \ding{55} & \ding{51} & \ding{51} \\
DaC \cite{Noohdani_2024_CVPR} & \ding{55} & Training & \ding{55} & \ding{51} & \ding{51} \\
\midrule
DISC \cite{wu23disc} & \ding{55} & Training & \ding{55} & \ding{51} & \ding{55} \\
CnC \cite{zhang2022correct} & \ding{55} & Training & \ding{55} & \ding{51} & \ding{55} \\
AFR \cite{qiu2023simple} & \ding{55} & Post-hoc & \ding{55} & \ding{51} & \ding{55} \\
EIIL \cite{creager2021environment} & \ding{55} & Training & \ding{55} & \ding{51} & \ding{55} \\
JTT \cite{liu2021just} & \ding{55} & Training & \ding{55} & \ding{51} & \ding{55} \\
\midrule
Post-hoc CBM \cite{yuksekgonul2023posthoc} & \ding{55} & Post-hoc & \ding{51} & \ding{51} & \ding{55} \\
Label-Free-CBM \cite{oikarinenlabel} & \ding{55} & Training & \ding{51} & \ding{55} & \ding{55} \\
LaBo \cite{yang2023language} & \ding{55} & Training & \ding{51} & \ding{55} & \ding{55} \\
CDM \cite{panousis2023sparse} & \ding{55} & Training & \ding{51} & \ding{55} & \ding{55} \\
DCLIP \cite{menon2022visual} & \ding{55} & Training & \ding{51} & \ding{55} & \ding{55} \\
DN\text{-}CBM \cite{rao2024discover} & \ding{55} & Training & \ding{51} & \ding{55} & \ding{55} \\
\midrule
SALF\text{-}CBM \cite{benou2025show} & \ding{51} & Training & \ding{51} & \ding{55} & \ding{55} \\
DCBM \cite{dcbm2025} & \ding{51} & Training & \ding{51} & \ding{55} & \ding{55} \\
\rowcolor[HTML]{C6EAD8} SEG-MIL-CBM (ours) & \ding{51} & Training & \ding{51} & \ding{51} & \ding{55} \\
\bottomrule
\end{tabular}
}
\caption{
    Full comparison of benchmark methods across multiple criteria. Grouped (top to bottom): methods with group annotations, methods without group annotations, non-spatially-aware CBMs, and spatially-aware CBMs. The trimmed CBM-only version appears as Table~1 in the main paper.
}
\label{apptab:comparison_criteria_full}
\end{table}

\begin{table}[h]
\centering
\footnotesize
\setlength{\tabcolsep}{0pt}
\begin{tabular*}{\columnwidth}{@{\extracolsep{\fill}}lcccc@{}}
\toprule
\textbf{Model} & \multicolumn{2}{c}{\textbf{Waterbirds (\%)}} & \multicolumn{2}{c}{\textbf{Pawrious (\%)}} \\
 & Avg. & Worst & Avg. & Worst \\
\midrule
ERM & 97.3 & 60.0 & 98.59 & 75.55 \\
\midrule
GroupDRO \cite{sagawa2019distributionally} & \underline{96.0} & 86.0 & 90.83 & \underline{86.67} \\
DFR \cite{kirichenko2022last} & 94.2 & \textbf{92.9} & -- & -- \\
DaC \cite{Noohdani_2024_CVPR} & 95.3 & \underline{92.3} & -- & -- \\
\midrule
CnC \cite{zhang2022correct} & 90.9 & 88.5 & -- & -- \\
AFR \cite{qiu2023simple} & 94.2 & 90.4 & \underline{98.77} & 82.22 \\
EIIL \cite{creager2021environment} & \textbf{96.9} & 78.7 & -- & -- \\
JTT \cite{liu2021just} & 93.6 & 86.7 & 98.45 & 82.26 \\
\midrule
\rowcolor[HTML]{C6EAD8} SEG-MIL-CBM (ours) & 85.73$\pm$0.18 & 72.00$\pm$1.08 & \textbf{97.73$\pm$0.01} & \textbf{87.41$\pm$0.20} \\
\bottomrule
\end{tabular*}
\caption{
    Group-robust training baselines: Accuracy on Waterbirds and Pawrious datasets. Avg: Average accuracy, Worst: Worst-group accuracy. SEG-MIL-CBM is included for reference; for Waterbirds, group labels are used for validation selection and evaluation, but not in the training loss. Legend: \textbf{bold} = best; \underline{underline} = second best.
}
\label{apptab:spurious_eval_dnn}
\end{table}

\clearpage
\section{Concept-Attention Bottleneck Ablation}
\label{appendix:concept_attention}

To test whether class-conditioned attention must read unconstrained segment embeddings, we train a Waterbirds variant in which the attention MLP receives learned concept activations $\myvec{z}_i$ instead of segment features $\myvec{h}_i$:
\[
u_{i,y}=\frac{\myvec{a}_y^\top\tanh(A_z\myvec{z}_i+\myvec{d})+e_y}{\tau_{\mathrm{attn}}},
\qquad
\alpha_{i,y}=\softmax_i(u_{i,y}).
\]
All other settings match the ViT-B/16 Waterbirds component ablation in the main paper: concept-guided segment cache, class-conditioned attention, $\lconcept=0.1$, area exponent $\gamma_{\mathrm{area}}=0.25$, and seeds $7200$--$7202$. The result is accuracy-neutral relative to the embedding-attention model, supporting the view that the reported Waterbirds behavior does not require an attention path around the learned concept activations.

\begin{table}[h]
\centering
\scriptsize
\resizebox{\columnwidth}{!}{%
\begin{tabular}{lcccccc}
\toprule
\textbf{Seed} & \textbf{Avg.} & \textbf{Worst} & \textbf{G0} & \textbf{G1} & \textbf{G2} & \textbf{G3} \\
\midrule
7200 & 85.16 & 71.18 & 96.11 & 71.18 & 72.86 & 98.31 \\
7201 & 85.47 & 72.90 & 96.57 & 73.52 & 72.90 & 98.27 \\
7202 & 85.43 & 72.43 & 96.11 & 72.43 & 73.26 & 98.27 \\
\midrule
\textbf{Mean $\pm$ std} & \textbf{85.35$\pm$0.17} & \textbf{72.17$\pm$0.89} & -- & -- & -- & -- \\
\bottomrule
\end{tabular}%
}
\caption{
    \textbf{Concept-attention Waterbirds ablation.} Attention is computed from learned concept activations $\myvec{z}_i$ rather than raw segment embeddings $\myvec{h}_i$. Group columns report Waterbirds group accuracies in percent.
}
\label{apptab:concept_attention_waterbirds}
\end{table}

\section{Waterbirds Foreground/Background Diagnostic}
\label{appendix:foreground_background}

We also evaluate whether the trained Waterbirds model can succeed from background evidence alone. For a random subset of $1000$ Waterbirds test images, we construct the union of retained segment masks from the cached preprocessing metadata. We then embed two single-instance diagnostic inputs with the same backbone: a foreground-union image that keeps pixels inside this mask union and fills the complement with the image mean, and a background-complement image that keeps pixels outside the union and fills the masked region. No model weights are updated. This is a mechanism diagnostic rather than a replacement benchmark, because the foreground union comes from the model's retained masks.

\begin{table}[h]
\centering
\scriptsize
\resizebox{\columnwidth}{!}{%
\begin{tabular}{lcccccc}
\toprule
\textbf{Variant} & \textbf{Avg.} & \textbf{Worst} & \textbf{G0} & \textbf{G1} & \textbf{G2} & \textbf{G3} \\
\midrule
Full cached bag & 84.7 & 70.5 & 97.6 & 70.5 & 77.0 & 98.2 \\
Foreground union only & 84.1 & 78.1 & 88.7 & 78.1 & 84.1 & 89.3 \\
Background complement only & 61.3 & 30.3 & 90.6 & 30.3 & 34.5 & 97.3 \\
\bottomrule
\end{tabular}%
}
\caption{
    \textbf{Waterbirds foreground/background diagnostic on $1000$ test images.} The foreground-union variant preserves most average accuracy and improves worst-group accuracy relative to the full cached bag on this subset, while the background-complement variant drops sharply.
}
\label{apptab:foreground_background_waterbirds}
\end{table}

\clearpage
\section{Attention-Only Faithfulness Control}
\label{appendix:attention_only_faithfulness}

We additionally evaluate a ranking that orders segments using only the predicted-class MIL attention weights, without multiplying by the class-conditioned concept evidence term. This control isolates the region-selection component of SEG-MIL-CBM's explanation under the same deletion/insertion protocol used in the main paper.

\begin{table}[h]
\centering
\scriptsize
\begin{tabular}{lcccc}
\toprule
\textbf{Ranking} & \textbf{DAUC} $\downarrow$ & \textbf{IAUC} $\uparrow$ & \textbf{Top-1 Ins.} $\uparrow$ & \textbf{Top-5 Del.} $\uparrow$ \\
\midrule
Attention only & 0.445 & 0.670 & 90.5 & 22.2 \\
SEG-MIL-CBM importance & 0.444 & 0.670 & 90.5 & 22.2 \\
\bottomrule
\end{tabular}
\caption{
    \textbf{Attention-only faithfulness control on CUB.} Attention-only ranking nearly matches the full contribution score under deletion/insertion, showing that the perturbation protocol mainly tests faithful region selection. The semantic value of the concept-conditioned term is evaluated separately in Appendix~\ref{appendix:cub_attribute_agreement}.
}
\label{apptab:attention_only_faithfulness}
\end{table}

\section{CUB Attribute Agreement}
\label{appendix:cub_attribute_agreement}

The deletion/insertion experiment in the main paper evaluates whether selected regions are faithful to the model prediction. It does not, by itself, test whether the concept labels attached to those regions are semantically correct. We therefore add a concept-level sanity check using CUB image attributes.

We keep the trained SEG-MIL-CBM concept vocabulary fixed and use no CUB attributes during training. For evaluation only, we map free-form model concepts to CUB attributes with a fixed lexical visual-attribute mapper. A concept is mapped only when the relevant visual part, pattern, shape, or color value matches the CUB attribute schema. This maps $279/369$ model concepts to at least one CUB attribute and covers $135/312$ CUB attributes. We then aggregate per-image concept scores into attribute scores and compare the resulting ranking against image-level CUB attributes with certainty at least $3$. Images with no positive attributes under this criterion are skipped, leaving $2913$ evaluated CUB test images. We use mAP as the primary summary because CUB attribute prediction is multi-label and each image may have many valid attributes; top-$k$ precision and recall are reported for completeness.

\begin{table}[h]
\centering
\scriptsize
\resizebox{\columnwidth}{!}{%
\begin{tabular}{lccccc}
\toprule
\textbf{Ranking signal} & \textbf{P@5} & \textbf{P@10} & \textbf{P@20} & \textbf{R@20} & \textbf{mAP} \\
\midrule
Random attributes & 9.7 & 9.7 & 9.7 & 6.4 & 11.2 \\
Attention-attached names & \textbf{29.9} & 24.6 & \textbf{22.8} & \textbf{15.3} & 17.1 \\
Activation only & 25.8 & 23.4 & 21.2 & 14.1 & 17.9 \\
\rowcolor[HTML]{C6EAD8} SEG-MIL-CBM contribution & 26.3 & \textbf{24.8} & 22.1 & 14.9 & \textbf{18.4} \\
\bottomrule
\end{tabular}%
}
\caption{
    \textbf{CUB attribute-agreement sanity check.} We map SEG-MIL-CBM's fixed free-form concept vocabulary to CUB attributes only for evaluation, then compare ranked concept-derived attribute scores against image-level CUB attribute annotations. Values are percentages over $2913$ evaluated CUB test images. The SEG-MIL-CBM contribution score improves mAP over the attention-attached-name control, supporting the semantic role of the concept-conditioned classifier term, while attention-attached names remain competitive on small top-$k$ metrics.
}
\label{apptab:cub_attribute_agreement}
\end{table}

As a stricter robustness check, we also repeat the evaluation using only attributes marked definitely present (certainty at least $4$), leaving $2620$ evaluated images. The qualitative pattern is unchanged: SEG-MIL-CBM contribution improves mAP over attention-attached names ($14.7$ vs.\ $14.0$), while attention-attached names remain stronger at the smallest top-$k$ values. We therefore interpret this diagnostic as evidence for semantic prioritization rather than as a universal improvement on every ranking metric.

\section{CIFAR-10-C Results}
\label{appendix:CIFAR10-C_results}

This section provides a corruption-robustness diagnostic on CIFAR-10-C. We report per-corruption error averaged over severities, severity-conditioned accuracy curves, and accuracy trends across corruption types; because performance varies by corruption, these results should be read as supplementary evidence rather than a single global robustness claim.

\begin{figure}[h]
    \centering
    \includegraphics[width=\linewidth]{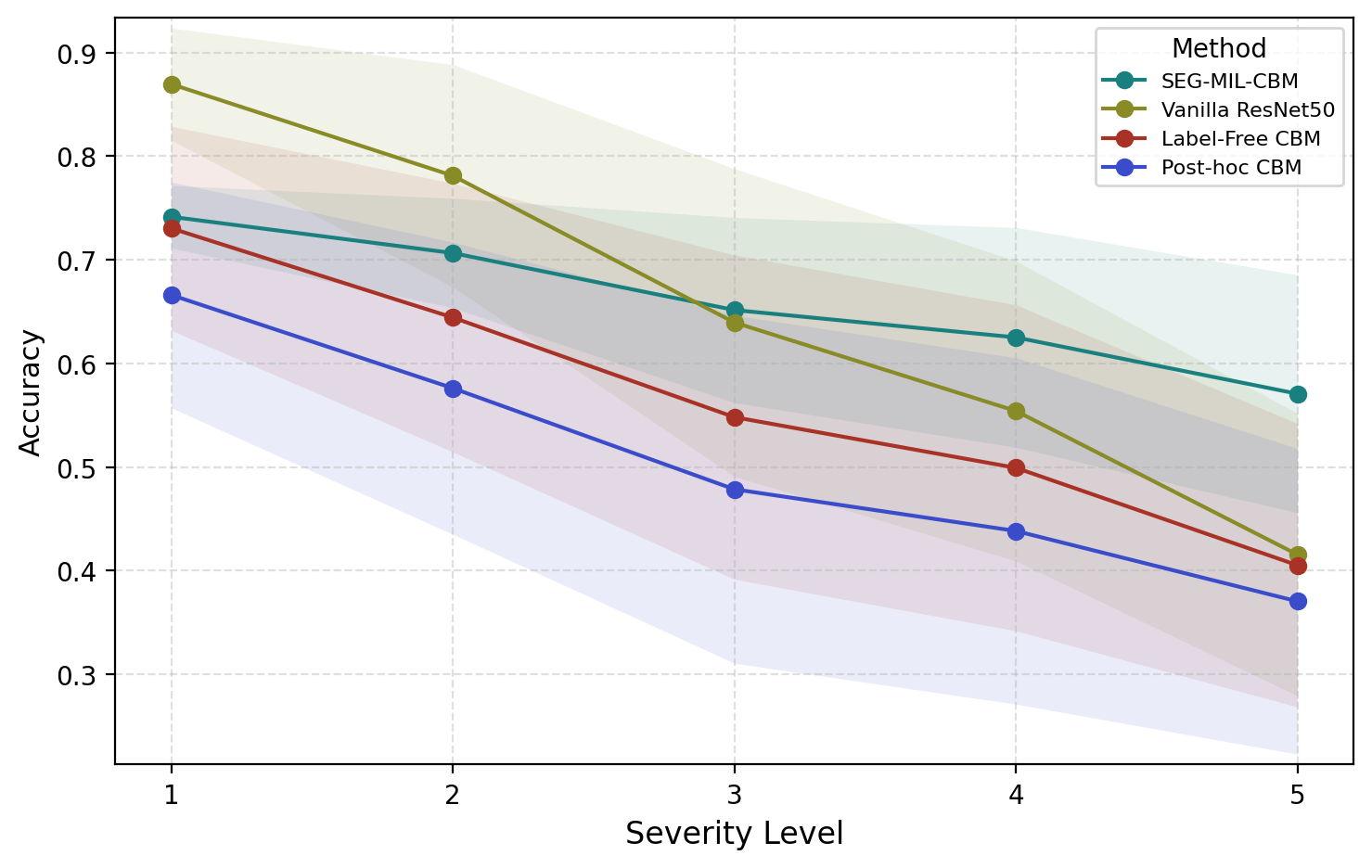}
    \caption{
        95\% confidence interval accuracy across 5 CIFAR-10-C corruptions (frost, gaussian blur, gaussian noise, shot noise, zoom blur) for: Vanilla ResNet-50 (pretrained on CIFAR-10), SEG-MIL-CBM, Label-Free CBM \cite{oikarinenlabel}, and Post-hoc CBM \cite{yuksekgonul2023posthoc}. In this five-corruption subset, SEG-MIL-CBM degrades less than baselines at higher severities (3--5), suggesting that segment-level aggregation can improve robustness for some common corruptions.
    }
    \label{appfig:cifar10c}
\end{figure}

\begin{table*}[t]
\centering
\caption{
    Per-corruption mCE (CE, lower is better) averaged over severities on CIFAR-10-C.
}
\label{apptab:mce_per_corruption}
\resizebox{\textwidth}{!}{
\begin{tabular}{lccccccccccccccc}
\toprule
\textbf{Method} & \textbf{gauss\_noise} & \textbf{shot\_noise} & \textbf{impulse\_noise} & \textbf{defocus\_blur} & \textbf{glass\_blur} & \textbf{motion\_blur} & \textbf{zoom\_blur} & \textbf{snow} & \textbf{frost} & \textbf{fog} & \textbf{brightness} & \textbf{contrast} & \textbf{elastic\_transf.} & \textbf{pixelate} & \textbf{jpeg\_comp.} \\
\midrule
\textbf{SEG-MIL-CBM} & 0.459 & 0.432 & 0.551 & 0.271 & 0.516 & 0.315 & 0.282 & 0.352 & 0.289 & 0.254 & 0.240 & 0.257 & 0.317 & 0.484 & 0.345 \\
\textbf{Label-Free-CBM (CE)} & 0.628 & 0.552 & 0.565 & 0.285 & 0.634 & 0.411 & 0.342 & 0.291 & 0.307 & 0.254 & 0.174 & 0.335 & 0.352 & 0.407 & 0.358 \\
\textbf{Post-hoc-CBM (CE)} & 0.796 & 0.721 & 0.693 & 0.420 & 0.773 & 0.587 & 0.472 & 0.393 & 0.424 & 0.348 & 0.321 & 0.459 & 0.546 & 0.562 & 0.575 \\
\bottomrule
\end{tabular}}
\end{table*}

\begin{figure*}[t]
    \centering
    \includegraphics[width=0.8\textwidth]{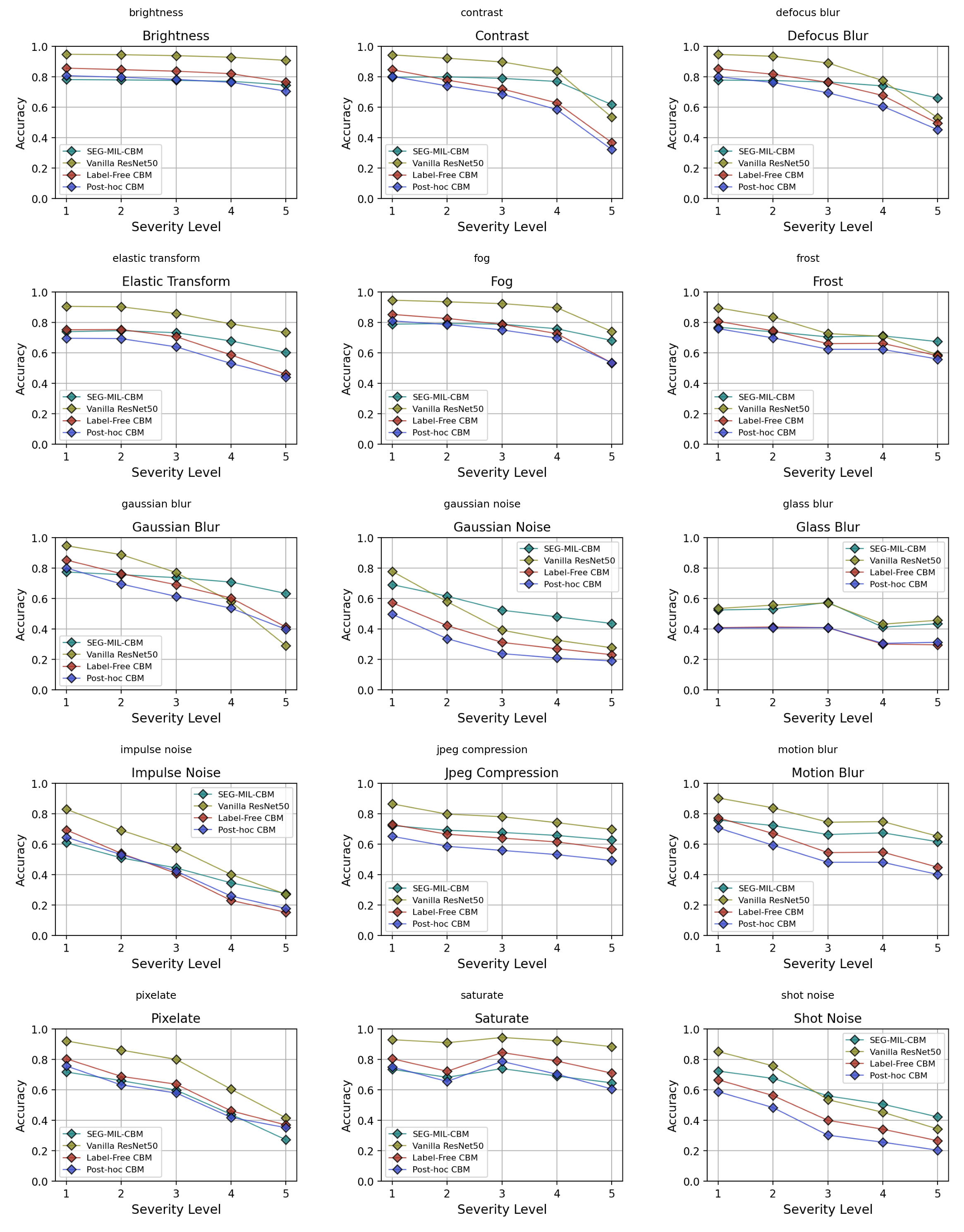}
    
    \caption{
        Accuracy trends across corruption types (page 1)
    }
    \label{fig:plots_per_corruption_grid_page001}
\end{figure*}

\begin{figure*}[t]
    \centering
    \includegraphics[width=1.0\textwidth,
          trim={0cm 23.0cm 0cm 0cm}, clip]{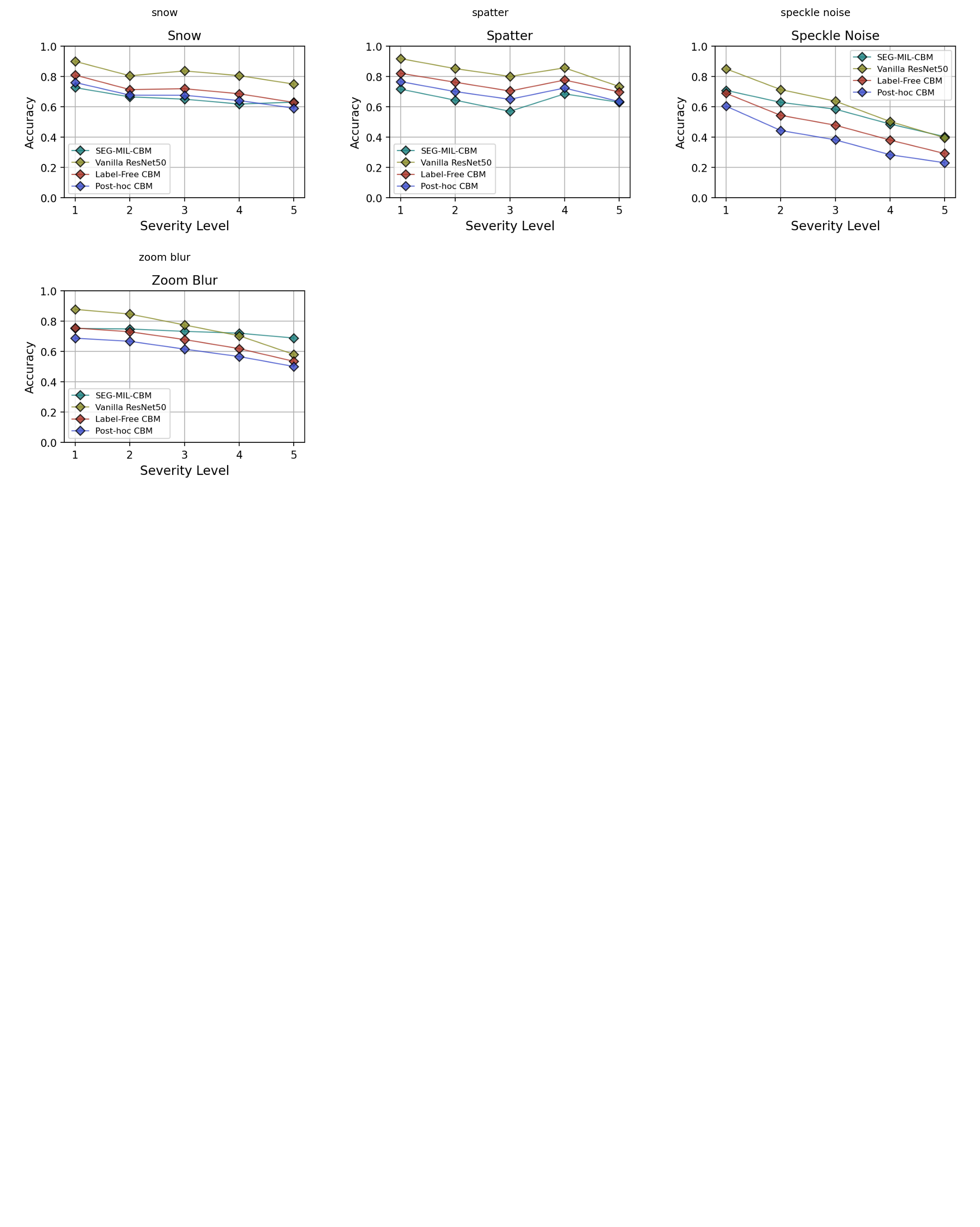}
    
    \caption{
        Accuracy trends across corruption types (page 2)
    }
    \label{fig:plots_per_corruption_grid_page002}
\end{figure*}

\clearpage
\section{Full Result Tables and Summary Statistics}
\label{appendix:full_per_seed_statistics}

This section collects the full result tables and summary statistics for datasets and backbones considered in the main paper. These tables complement the abbreviated main-text results and report mean and standard deviation where multiple runs are available.

\begin{table}[h]
\centering

\scriptsize
\resizebox{\columnwidth}{!}{
\begin{tabular}{lccccc}
\toprule
 & IMN & Places & CUB & CIFAR-10 & CIFAR-100 \\
\midrule
Linear Probe \cite{dcbm2025} & 80.2 & 55.1 & 81.0 & 96.2 & 86.4 \\
Zero Shot \cite{dcbm2025} & 68.6 & 39.5 & 55.0 & 91.6 & 68.7 \\
\midrule
Label-Free-CBM \cite{oikarinenlabel,dcbm2025} & 75.4 & 48.2 & 74.0 & 94.7 & 77.4 \\
Post-hoc-CBM \cite{yuksekgonul2023posthoc} & -- & -- & 61.0 & 87.1 & 68.0 \\
LaBo \cite{yang2023language} & 78.9 & -- & -- & \underline{95.7} & 81.2 \\
CDM \cite{panousis2023sparse} & 79.3 & \underline{52.6} & \textbf{79.5} & 95.3 & \underline{80.5} \\
DCLIP \cite{menon2022visual} & 68.0 & 40.3 & 57.8 & -- & -- \\
DN-CBM \cite{rao2024discover} & \textbf{79.5} & \textbf{55.1} & -- & \textbf{96.0} & 82.1 \\
\midrule
DCBM-SAM2 \cite{dcbm2025} & 70.4 & 50.6 & 75.3 & 95.2 & 79.4 \\
DCBM-GDINO \cite{dcbm2025} & 69.7 & 50.7 & 74.1 & 95.1 & 79.6 \\
DCBM-MASKRCNN \cite{dcbm2025} & 70.5 & 50.9 & 76.7 & 95.2 & 79.6 \\
SALF-CBM \cite{benou2025show} & \underline{78.6} & 49.4 & 76.2 & -- & -- \\
\rowcolor[HTML]{C6EAD8} SEG-MIL-CBM (ours) & 78.5$\pm$0.2 & 51.24$\pm$0.25 & \underline{77.39$\pm$0.22} & 94.89$\pm$0.12 & \textbf{85.26$\pm$0.00} \\
\bottomrule
\end{tabular}
}
\caption{
    Accuracy with CLIP ViT-B/16 backbone on ImageNet (IMN), Places, CUB, CIFAR-10, and CIFAR-100.
}
\label{apptab:clip_vitb16_full_stats}
\end{table}
\begin{table}[h]
\centering

\footnotesize
\setlength{\tabcolsep}{0pt}
\begin{tabular*}{\columnwidth}{@{\extracolsep{\fill}}lcccc@{}}
\toprule
\textbf{Model} & \multicolumn{2}{c}{\textbf{Waterbirds (\%)}} & \multicolumn{2}{c}{\textbf{Pawrious (\%)}} \\
 & Avg. & Worst & Avg. & Worst \\
\midrule
ERM & 97.3 & 60.0 & 98.59 & 75.55 \\
\midrule
Label-Free-CBM \cite{oikarinenlabel} & \underline{81.82$\pm$0.01} & 54.62$\pm$0.00 & 94.67$\pm$0.002 & \underline{46.67$\pm$0.02} \\
Post-hoc-CBM \cite{yuksekgonul2023posthoc} & 80.58$\pm$0.07 & \underline{57.89$\pm$1.72} & 91.20$\pm$0.26 & 19.26$\pm$2.57 \\
\midrule
SALF-CBM$^\dagger$ \cite{benou2025show} & 78.89 & 65.11 & -- & -- \\
\midrule
\rowcolor[HTML]{C6EAD8} SEG-MIL-CBM (ours) & \textbf{85.73$\pm$0.18} & \textbf{72.00$\pm$1.08} & \textbf{97.73$\pm$0.01} & \textbf{87.41$\pm$0.20} \\
\bottomrule
\end{tabular*}
\caption{
    CBM baselines: Accuracy on Waterbirds, and Pawrious datasets. Avg: Average accuracy, Worst: Worst-group accuracy. For Waterbirds, SEG-MIL-CBM reports a fixed configuration averaged over five seeds; group labels are used for validation selection and evaluation, but not in the training loss. $^\dagger$ denotes our local SALF-style Waterbirds run under the closest matching ViT-B/16 protocol, included because full SALF-CBM training code was not publicly available at the time of our experiment.
}
\label{apptab:spurious_eval_cbm}
\end{table}

\begin{table}[h]
\centering

\footnotesize
\setlength{\tabcolsep}{1pt}
\begin{tabular*}{\columnwidth}{@{\extracolsep{\fill}}lcccc@{}}
\toprule
\textbf{Model} & \multicolumn{2}{c}{\textbf{Waterbirds (\%)}} & \multicolumn{2}{c}{\textbf{Pawrious (\%)}} \\
 & Avg. & Worst & Avg. & Worst \\
\midrule
ERM & 97.3 & 60.0 & 98.59 & 75.55 \\
\midrule
GroupDRO \cite{sagawa2019distributionally} & \underline{96} & 86.0 & 90.83 & \underline{86.67} \\
DFR \cite{kirichenko2022last} & 94.2$\pm$0.4 & \textbf{92.9$\pm$0.2} & -- & -- \\
DaC \cite{Noohdani_2024_CVPR} & 95.3$\pm$0.4 & \underline{92.3$\pm$0.4} & -- & -- \\
\midrule
CnC \cite{zhang2022correct} & 90.9$\pm$0.1 & 88.5$\pm$0.3 & -- & -- \\
AFR \cite{qiu2023simple} & 94.2$\pm$1.2 & 90.4$\pm$1.1 & \underline{98.77} & 82.22 \\
EIIL \cite{creager2021environment} & \textbf{96.9} & 78.7 & -- & -- \\
JTT \cite{liu2021just} & 93.6 & 86.7 & 98.45 & 82.26 \\
\midrule
\rowcolor[HTML]{C6EAD8} SEG-MIL-CBM (ours) & 85.73$\pm$0.18 & 72.00$\pm$1.08 & \textbf{97.73$\pm$0.01} & \textbf{87.41$\pm$0.20} \\
\bottomrule
\end{tabular*}
\caption{
    Group-robust training baselines: Accuracy on Waterbirds and Pawrious datasets. Avg: Average accuracy, Worst: Worst-group accuracy. For Waterbirds, SEG-MIL-CBM uses group labels for validation selection and evaluation, but not in the training loss.
}
\label{apptab:spurious_eval_dnn_full_stats}
\end{table}

\begin{table}[h]
\centering
\scriptsize
\resizebox{\columnwidth}{!}{
\begin{tabular}{lccccc}
\toprule
 & IMN & Places & CUB & CIFAR-10 & CIFAR-100 \\
\midrule
Linear Probe \cite{dcbm2025} & 73.3 & 53.4 & 68.9 & 88.7 & 76.3 \\
Zero Shot \cite{dcbm2025} & 59.6 & 37.9 & 46.1 & 75.6 & 41.6 \\
\midrule
Label-Free-CBM \cite{oikarinenlabel} & 72.0 & 46.8 & 74.3 & 86.4 & 65.1 \\
Post-hoc-CBM \cite{yuksekgonul2023posthoc} & -- & -- & 61.0 & 87.1 & 68.0 \\
LaBo \cite{yang2023language} & 68.9 & -- & -- & 87.9 & 69.1 \\
CDM \cite{panousis2023sparse} & 72.2 & 52.7 & 72.3 & 86.5 & 67.6 \\
DCLIP \cite{menon2022visual} & 59.6 & 37.9 & 49.0 & -- & -- \\
DN-CBM \cite{rao2024discover} & 72.9 & 53.5 & -- & 87.6 & 67.5 \\
\midrule
DCBM-SAM2 \cite{dcbm2025} & 58.7 & 48.0 & 61.4 & 84.5 & 61.8 \\
DCBM-GDINO \cite{dcbm2025} & 58.7 & 47.8 & 59.0 & 83.9 & 61.2 \\
DCBM-MASKRCNN \cite{dcbm2025} & 58.7 & 48.2 & 64.6 & 84.5 & 62.7 \\
SALF-CBM (Sparse) \cite{benou2025show} & 75.32 & 46.73 & 74.35 & -- & -- \\
SALF-CBM \cite{benou2025show} & 76.26 & 49.38 & 76.21 & -- & -- \\
\midrule
\rowcolor[HTML]{C6EAD8} SEG-MIL-CBM (ours) & 76.02 & 48.05 & 76.79 & 89.8 & 76.71 \\
\bottomrule
\end{tabular}
}
\caption{
    Accuracy with CLIP ResNet-50 backbone on ImageNet (IMN), Places, CUB, CIFAR-10, and CIFAR-100.
}
\label{apptab:clip_r50_only}
\end{table}

\clearpage
\section{Sensitivity Analysis}
\label{app:sensitivity}

This section provides supplementary analysis of SEG-MIL-CBM's sensitivity to key hyperparameters used in our implementation. In particular, we examine the concept-alignment weight $\lambda_{\text{concept}}$ and the number of segments $N_s$.

\begin{figure}[h]
    \centering
    \includegraphics[width=\linewidth]{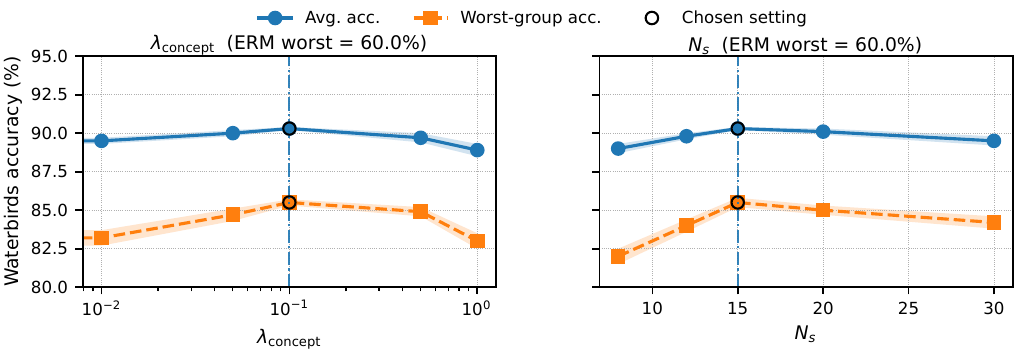}
    \caption{
        \textbf{Sensitivity analysis on Waterbirds.} Average and worst-group accuracy across (left) concept-alignment weight $\lambda_{\text{concept}}$ (log scale) and (right) number of retained segments $N_s$. Hollow markers and vertical dash–dot lines indicate the default settings used throughout all experiments.
    }
    \label{fig:sensitivity_waterbirds}
\end{figure}

Figure~\ref{fig:sensitivity_waterbirds} reports average accuracy and worst-group accuracy on the \textbf{Waterbirds} benchmark across a range of values for both hyperparameters.

\clearpage
\section{Implementation Details.}
\label{app:impl}

This section summarizes the implementation details used across datasets in our experiments. We list the hyperparameters for concept selection, mask filtering, bag construction, and training, providing a reproducible configuration for SEG-MIL-CBM.

\begin{table}[h]
\centering
\scriptsize
\caption{
    \textbf{Implementation details per dataset.} Hyperparameters for concept selection, mask filtering, bag construction, and training. $\tauminpix$ = minimum mask size (in pixels); $\rho_{\max}$ = maximum mask area ratio; $N_s$ = bag size; $D$ = backbone feature dimension.
}
\label{tab:impl_per_dataset}
\resizebox{\columnwidth}{!}{%
\begin{tabular}{lccccc}
\toprule
\textbf{Setting} 
& \textbf{Waterbirds} 
& \textbf{Pawrious} 
& \textbf{CIFAR-10/100} 
& \textbf{CUB} 
& \textbf{IMN / Places} \\
\midrule

Top-$K$ concepts $K_{\text{top}}$      
& 32 & 32 & 32 & 32 & 64 \\

Image backbone                         
& ResNet-50 & ResNet-50 & ResNet-50 & ResNet-50 & ResNet-50 / ViT-B/16 \\

Feature dim.\ $D$                      
& 2048 & 2048 & 2048 & 2048 & 2048 / 768 \\

CLIP backbone                          
& ViT-B/32 / ViT-B/16 & ViT-B/32 & ViT-B/16 & ViT-B/16 & ViT-B/16 \\

Mask size thresholds                   
& \multicolumn{5}{c}{$\tauminpix$, $\rho_{\max}{=}0.5$} \\

IoU merge threshold $\tauIoU$          
& \multicolumn{5}{c}{$\tauIoU{=}0.5$} \\

Bag size $N_s$                         
& 15 & 15 & 5 & 15 & 5 \\

Concept loss weight $\lconcept$        
& \multicolumn{5}{c}{$\lconcept{=}0.1$} \\

Attention module                       
& \multicolumn{5}{c}{Class-conditioned MLP (hidden 128, \textit{Tanh}), softmax over segments per class} \\

Optimizer / LR                         
& \multicolumn{5}{c}{Adam, $1\times10^{-4}$} \\

Random seeds                           
& \multicolumn{5}{c}{3 runs unless noted; Waterbirds headline uses 5 seeds} \\

\bottomrule
\end{tabular}%
}
\end{table}

\end{document}